\documentclass[review]{elsarticle}

\usepackage[table,xcdraw]{xcolor}

\usepackage{graphicx}
\usepackage{amsmath}
\usepackage{natbib}
\usepackage{eurosym}
\usepackage{graphicx}
\usepackage{wrapfig}
\usepackage[english]{babel}
\usepackage[center]{caption}
\usepackage{mathtools}
\usepackage{footnote}
\makesavenoteenv{tabular}
\makesavenoteenv{table}
\usepackage{latexsym}
\usepackage{url}
\usepackage{tikz}
\usepackage[amsmath,thmmarks]{ntheorem}
\usepackage{amsmath}
\usepackage{amssymb}
\usepackage{amsmath}

\usepackage{multirow}
\usepackage{capt-of, blindtext}
\usepackage{pdfpages}
\usepackage{setspace}
\usepackage{algorithmic}
\usepackage{algorithm}
\usepackage{fullpage}
\usepackage{kpfonts}
\usepackage{setspace}
\usepackage{algorithmic}
\usepackage{algorithm}
\usepackage{mdframed}
\usepackage{bbold}
\usepackage{multicol}
\usepackage{hyperref}

\usepackage[table]{xcolor}
\usepackage{xcolor}

\usepackage{microtype}

\mdfdefinestyle{MyFrame}{%
                linecolor=black,
                outerlinewidth=2pt,
                roundcorner=10pt,
                innertopmargin=0.3cm,
                innerbottommargin=0.3cm,
                innerrightmargin=10pt,
                innerleftmargin=10pt,
                backgroundcolor=white}%gray!20!white}

\definecolor{mygreen}{RGB}{14, 107, 14}
\definecolor{myred}{RGB}{151, 0, 0}

\newcommand{\greencheck}{}%
\DeclareRobustCommand{\greencheck}{%
  \tikz\fill[scale=0.65, color=mygreen]
  (0,.35) -- (.25,0) -- (1,.7) -- (.25,.15) -- cycle;%
}

\newcommand{\crosscheck}{$\mathbin{\tikz [x=2.4ex,y=2.4ex,line width=.4ex, myred] \draw (0,0) -- (1,1) (0,1) -- (1,0);}$}%

\journal{Preprint}

\begin{document}

\begin{frontmatter}

\title{Counterfactuals and Causability in Explainable Artificial Intelligence: Theory, Algorithms, and Applications}
%\tnotetext[mytitlenote]{Fully documented templates are available in the elsarticle package on \href{http://www.ctan.org/tex-archive/macros/latex/contrib/elsarticle}{CTAN}.}

%% Group authors per affiliation:
\author{Yu-Liang Chou$^1$, Catarina Moreira\corref{mycorrespondingauthor}$^{1,2}$, Peter Bruza$^1$, Chun Ouyang$^1$, Joaquim Jorge$^2$}
\address{$^1$ School of Information Systems, Queensland University of Technology, Brisbane, Australia}
\address{$^2$ INESC-ID Lisboa, Instituto Superior T\'{e}cnico, ULisboa, Portugal}

\begin{abstract} 

Deep learning models have achieved high performances across different domains, such as medical decision-making, autonomous vehicles, decision support systems, etc. Despite this success, the internal mechanisms of these models are opaque because their internal representations are too complex for a human to understand. This makes it hard to understand the \textit{how} or the \textit{why} of the predictions of deep learning models.

There has been a growing interest in model-agnostic methods that can make deep learning models more transparent and explainable to a user. Some researchers recently argued that for a machine to achieve a certain degree of human-level explainability, this machine needs to provide human causally understandable explanations, also known as \textit{causability}. A specific class of algorithms that have the potential to provide causability are \textit{counterfactuals}.

This paper presents an in-depth systematic review of the diverse existing body of literature on counterfactuals and causability for explainable artificial intelligence. We performed an LDA topic modeling analysis under a PRISMA framework to find the most relevant literature articles. This analysis resulted in a novel taxonomy that considers the grounding theories of the surveyed algorithms, together with their underlying properties and applications in real-world data.  
This research suggests that current model-agnostic counterfactual algorithms for explainable AI are not grounded on a causal theoretical formalism and, consequently, cannot promote causability to a human decision-maker. Our findings suggest that the explanations derived from major algorithms in the literature provide spurious correlations rather than cause/effects relationships, leading to sub-optimal, erroneous, or even biased explanations. This paper also advances the literature with new directions and challenges on promoting causability in model-agnostic approaches for explainable artificial intelligence. 
\end{abstract}


\end{frontmatter}

% to cite
% Johansson16. Model specific, but nice paper

\section{Introduction}

% something to add
% While rule-based solutions of the early AI in the 1950s represented comprehensible “glass box” approaches, their weak- ness lay in dealing with uncertainties of the real world. Many problems from our everyday lives cannot be represented by for- mal, mathematical rules of logic. The failure of such algorithms to solve problems that are relatively simple for humans, such as natural language, recognizing faces, or understanding a joke, ultimately led to the “AI winter” in the 1980s.

% context - provide some context to orient those readers who are less familiar with your topic and to establish the importance of your work.

Artificial intelligence,in particular, deep learning, have made great strides in equaling, and even surpassing human performance in many tasks such as categorization, recommendation, game playing or even in medical decision-making~\citep{Moreira20medical}. Despite this success, the internal mechanisms of these technologies are an enigma because humans cannot scrutinize how these intelligent systems do what they do. This is known as the \textit{black-box} problem~\citep{Lipton2018}. Consequently, humans are reliant to blindly accept the answers produced by machine intelligence without understanding how that outcome came to be. There is growing disquiet about this state of affairs as intelligent technologies increasingly support human decision-makers in high-stakes contexts such as the battlefield, law courts, operating theatres, etc.

\subsection{The need for Explainability}

Several factors motivated the rise of approaches that attempt to turn predictive black-boxes transparent to the decision-maker~\citep{Doran2017,Mothila20criticize}. One of these factors is the recent European General Data Protection Regulation (GDPR)~\citep{Goodman2017}, which made the audit and verifiability of decisions from intelligent autonomous systems mandatory, increasing the demand for the ability to question and understand Machine Learning (ML) systems. These regulations directly impact worldwide businesses because GDPR applies not only to data being used by European organisations, but also to European data being used by other organisations. Another important factor is concerned with discrimination (such as gender and racial bias)~\citep{ONeil17}. Studies suggest that predictive algorithms widely used in healthcare, for instance, exhibited racial biases that prevented minority societal groups from receiving extra care~\citep{Obermeyer19} or display cognitive biases associated with medical decisions~\citep{Lau07,Saposnik16}. In medical X-ray images, it was found that deep learning models have learned to detect a metal token that technicians use to visualise the X-ray images, making this feature impacting the predictions of the algorithm~\citep{Zech2018}.  Other studies revealed gender and racial biases in automated facial analysis algorithms made available by commercial companies~\citep{Buolamwini18}, gender biases in textual predictive models~\citep{Bolukbasi16,Garg19,Caliskan17} or even more discriminatory topics such as facial features according to sexual orientation~\citep{Kosinski18}. 

The black-box problem and the need for interpretability motivated an extensive novel body of literature in machine learning that focuses on developing new algorithms and approaches that not only can interpret the complex internal mechanisms of machine learning predictions but also explain and provide understanding to the decision-maker of the $why$ these predictions~\citep{Lakkaraju2019,Doshivelez2017}. In this sense, interpretability and explainability have become the main driving pillars of explainable AI (XAI)~\citep{Gilpin18}. More specifically, we define interpretability and explainability in the following way.

\begin{itemize}
    \item \textbf{Interpretability} is defined as the extraction of relevant sub-symbolic information from a machine-learning model concerning relationships either contained in data or learned by the model~\citep{Murdoch19}. 
    
    \item \textbf{Explainability}, on the other hand, refers to the ability to translate this sub-symbolic information in a comprehensible manner through human-understandable language expressions~\citep{Holzinger19}.
\end{itemize}

The overarching goal of XAI is to generate human-understandable explanations of the \textit{why} and the \textit{how} of specific predictions from machine learning or deep learning (DL) systems.
\citet{Paez19} extends this goal by adding that explainable algorithms should offer a pragmatic and naturalistic account of understanding in AI predictive models and that explanatory strategies should offer well-defined goals when providing explanations to its stakeholders.

%Unsatisfactory XAI models can deliver unacceptable outcomes that can put minority groups at risk, such as providing erroneous medical diagnoses or yielding racially discriminatory decisions. 

Currently, there is an extensive body of literature reviewing different aspects of XAI

\textcolor{blue}{In \citet{Miller2019}, the author portraits the missing link between the current research on explanations from the fields of philosophy, psychology, and cognitive science. \citet{Miller2019} highlighted three main aspects that an XAI system must have in order to achieve explainability: (1) people seek explanations of the for \textit{why some event happened, instead of another?}, which suggests a need for counterfactual explanations; (2) recommendations can focus on a selective number of causes (not all of them), which suggests the need for causality in XAI; and (3) explanations should consist in conversations and interactions with a user promoting an explanation process where the user engages in and learns the explanations.} 
\textcolor{blue}{In \citet{Guidotti18survey}, the authors survey black-box specific methods for explainability and propose a taxonomy for XAI systems based on four features: (1) the type of problem faced based on their taxonomy; (2) the type of explainer adopted; (3) the type of black-box model that the explainer can process; and (4) the type of data that the black-box supports.}
\textcolor{blue}{On the other hand \citet{Das2020opportunities} proposed a taxonomy for categorizing XAI techniques based on their scope of explanations, methodology behind the algorithms,
and explanation level or usage. \citep{Adadi18} classified explainable methods according to (1) the interpretability of the model to be explained; (2) the scope of the interpretability; and (3) if the black-box is dependent or not on any type of machine learning model.} 

\textcolor{blue}{\citet{Arrieta2020} propose and discuss a taxonomy related to explainability in different machine learning and deep learning models. Additionally, the authors propose new methodologies towards responsible artificial intelligence and discuss several aspects regarding fairness, explainability, and accountability in real-world organisations.}

\textcolor{blue}{Some authors have reviewed evaluation methodologies for explainable systems and proposed a novel categorisation of XAI design goals and evaluation measures according to stakeholders~\citep{Mohseni20}. In contrast, other authors identified objectives that evaluation metrics should achieve and demonstrated the subjectiveness of evaluation measures regarding human-centered XAI systems~\citep{Zhou21metrics}. \citet{Carvalho19} extensively surveyed the XAI literature with a focus on both qualitative and quantitative evaluation metrics as well as properties/axioms that explanations should have. For other examples of works that survey evaluation in XAI, the reader can refer to \citet{hoffman2019metrics,AlvarezMelis2018}.}

\textcolor{blue}{In terms of the generation of explanations, \citet{Chen2020bias} surveyed the literature in terms of biases. The authors identified seven different types of biases that were found in recommendations and proposed a taxonomy in terms of current work on recommendations and potential ways to \textit{dibiase} them.}

\subsection{From Explainability to the need of Causability}

For a model to be interpretable, it must suggest explanations that make sense to the decision-maker and ensure that those explanations accurately represent the true reasons for the model's decisions~\cite{Serrano2019attention}. Current XAI models that attempt to decipher a black-box that is already trained (also known as post-hoc, model-agnostic models) build models around local interpretations, providing approximations to the predictive black-box~\citep{Ribeiro16,Lundberg17}, instead of reflecting the true underlying mechanisms of the black box (as pointed out by~\cite{Rudin2019}). In other words, these algorithms compute correlations between individual features to approximate the predictions of the black-box. In this paper, we argue that the inability to disentangle \textit{correlation from causation} can deliver sub-optimal or even erroneous explanations to decision-makers~\citep{Richens2020}. Causal approaches should be emphasized in XAI to promote a higher degree of interpretability to its users and avoid biases and discrimination in predictive black-boxes \citep{Kilbertus17}. 

The ability to find causal relationships between the features and the predictions in observational data is a challenging problem and constitutes a fundamental step towards explaining predictions~\citep{Holzinger19}. Causation is a ubiquitous notion in Humans' conception of their environment \citep{Pearl88}. Humans are extremely good at constructing mental decision models from very few data samples because people excel at generalising data and tend to think in cause/effect manners~\citep{Byrne19}. There has been a growing emphasis that AI systems should be able to build causal models of the world that support explanation and understanding, rather than merely solving pattern recognition problems~\cite{Lake17}. For decision-support systems, whether in finance, law, or even warfare, understanding the causality of learned representations is a crucial missing link~\cite{Pearl09,Gershman15,Peters17}. However, when considering machines, how can we make computer-generated explanations causally understandable by humans? This notion was recently put forward by~\citet{Holzinger19} in a term coined \textit{causability}.

\begin{figure}[h!]
    \resizebox{\columnwidth}{!} {
    \includegraphics{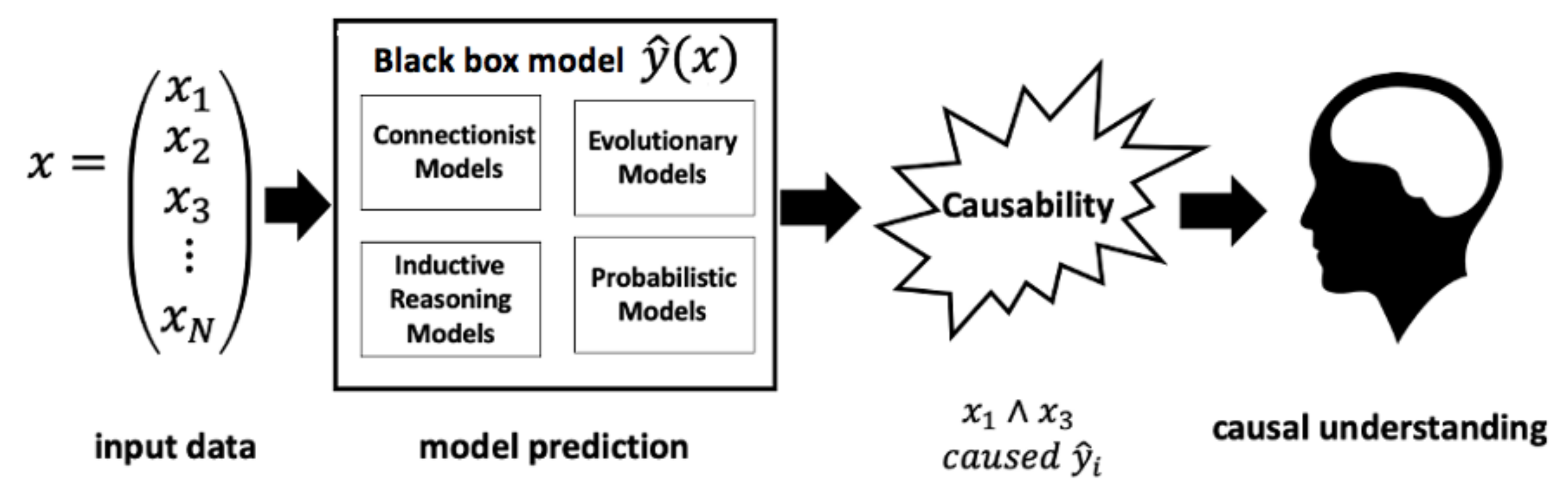}
    }
    \caption{The notion of causability within this thesis: given a predictive black box model, the goal is to create interpretable and explainable methods that will provide the user a causal understanding of why certain features contributed to a specific prediction~\citep{Holzinger19, Holzinger20icom,Holzinger21}}
    \label{fig:causability}
\end{figure}

    \begin{itemize}
        \item \textbf{Causability:} the extent to which an explanation of a statement to a human expert achieves a specified level of causal understanding with effectiveness, efficiency, and satisfaction in a specified context of use~\citep{Holzinger19}. In this sense, causability can be seen as a property of \textit{human intelligence}, whereas explainability as a property of \textit{artificial intelligence}~\citep{Holzinger20icom}.
        %It lends the justification for what and how should be explained as it determines the relative importance of the properties of explainability~\citep{Shin2021}.
        Figure~\ref{fig:causability} illustrates the notion of \textit{causability} under the context of XAI.
    \end{itemize}

\subsection{Counterfactuals as Means to Achieve Causability}

Causality is a fundamental concept to gain \textit{intellectual understanding} of the universe and its contents. It is concerned with establishing cause-effect relationships \citep{Hoque21}. Causal concepts are central to our practical deliberations, health diagnosis, etc. \citep{Mothilal20, Holzinger19}. Even when one attempts to explain certain phenomena, the explanation produced must acknowledge, to a certain degree, the causes of the effects being explained~\citep{Halpern05part1}. However, the nature and definition of causality is a topic that has promoted a lot of disagreement throughout the centuries in Philosophical literature. While Bertrand Russel was known for being the most famous denier of causality, arguing that it constituted an incoherent topic~\citep{Psillos02}, it was mainly with the philosopher and empiricist David Hume that the concept of causation started to be formally analyzed in terms of \textit{sufficient} and \textit{necessary conditions}: an event $c$ causes an event $e$ if and only if there are event-types $C$ and $E$ such that $C$ is necessary and sufficient for $E$~\citep{Psillos02}. Hume was also one of the first philosophers to identify causation through the notion of counterfactual: a cause to be an object followed by another in which if the first object (the cause) had not occurred, then the second (the effect) would never exist~\citep{Hume1739}. This concept started to gain more importance in the literature with the works of \citet{Lewis73}. 

Counterfactuals are then defined as a conditional assertion whose antecedent is false and whose consequent describes how the world would have been if the antecedent had occurred (a \textit{what-if} question). In the field of XAI, counterfactuals provide interpretations as a means to point out which changes would be necessary to accomplish the desired goal (prediction), rather than supporting the understanding of why the current situation had a certain predictive outcome \citep{Wachter18}. While most XAI approaches tend to focus on answering \textit{why} a certain outcome was predicted by a black-box, counterfactuals attempt to answer this question in another way by helping the user understand \textit{what features does the user need to change to achieve a certain outcome}~\citep{Poyiadzi20}. For instance, in a scenario where a machine learning algorithm assesses whether a person should be granted a loan or not, a counterfactual explanation of \textit{why} a person did not have a loan granted could be in a form of a scenario \textit{if your income was greater than $\$15,000$ you would be granted a loan} \citep{Mothilal20}.

\subsection{Contributions}

The hypothesis that we put forward in this paper is that the inability to disentangle \textit{correlation from causation} can deliver sub-optimal or even erroneous explanations to decision-makers~\citep{Richens2020}. Causal approaches should be emphasized in XAI to promote a higher degree of interpretability to its users. In other words, to achieve a certain degree of human-understandable explanations, causability should be a necessary condition. 

Given that there is not a clear understanding of the current state of the art concerning causal and causability approaches to XAI, it is the purpose of this paper to make a systematic review and critical discussion of the diverse existing body of literature on these topics. This systematic review will introduce researchers in the field of XAI that are interested in focusing their knowledge on the current state-of-the-art approaches currently present in the literature. Recently two papers surveying counterfactuals in XAI were proposed in the literature~\citep{Verma20,Stepin2021,Karimi2021survey}. Our paper distinguishes from the current body of literature by making a deep analysis on current counterfactual model-agnostic approaches, and how they could promote causability.

% contributions
In summary, this paper contributes to a literature review with discussions under three paradigms:

\begin{itemize}
    \item \textbf{Theory.} Survey and formalize the most important theoretical approaches that ground current explainable AI models in the literature that are based on counterfactual theories for causality.
    
    \item \textbf{Algorithms.} Understand what are the main algorithms that have been proposed in the XAI literature that use counterfactuals, and discuss which ones have are based on probabilistic approaches for causality and which ones have the potential to achieve a certain degree of causability.
    
    \item \textbf{Applications.} A continuous use case analysis to understand the main domains and fields where XAI algorithms that promote causability are emerging and the potential advantages and disadvantages of such approaches in real-world problems, namely in the mitigation of biased predictions.
\end{itemize}

\subsection{Paper Organisation}

This paper is organised as follows. In Section~\ref{sec:xai}, we present the taxonomy of the current state-of-the-art algorithms in XAI. In Section~\ref{sec:systematic_review} we present a systematic review on counterfactual and causability approaches in XAI. In the following sections, we present the findings of our systematic literature review.  Section~\ref{sec:properties}, we present properties that are used throughout the literature to assess what is a good counterfactual and a discussion on the impacts of different distance functions on XAI algorithms. Section~\ref{sec:counterfactuals_theory}, we present our first contribution where we analyse the theories that underpin the different algorithms of the literature by introducing a novel taxonomy for model-agnostic counterfactuals in XAI. In Section~\ref{sec:counterfactuals_algo}, we analyse the different algorithms of the literature based on the taxonomy that we proposed. In Section~\ref{sec:counterfactuals_applications}, we discuss the main applications of XAI algorithms together with recent developments in causability. In Section~\ref{sec:causability}, we present the main characteristics that should be part of a causability system for XAI. Finally, Section~\ref{sec:rq} answers the proposed research questions, and Section~\ref{sec:conclusion} presents the main conclusions of the work.

\section{A General Overview of Current Model-Agnostic Approaches in XAI}\label{sec:xai}

\begin{figure}[!h]
    \centering
    \resizebox{\columnwidth}{!} {
    \includegraphics{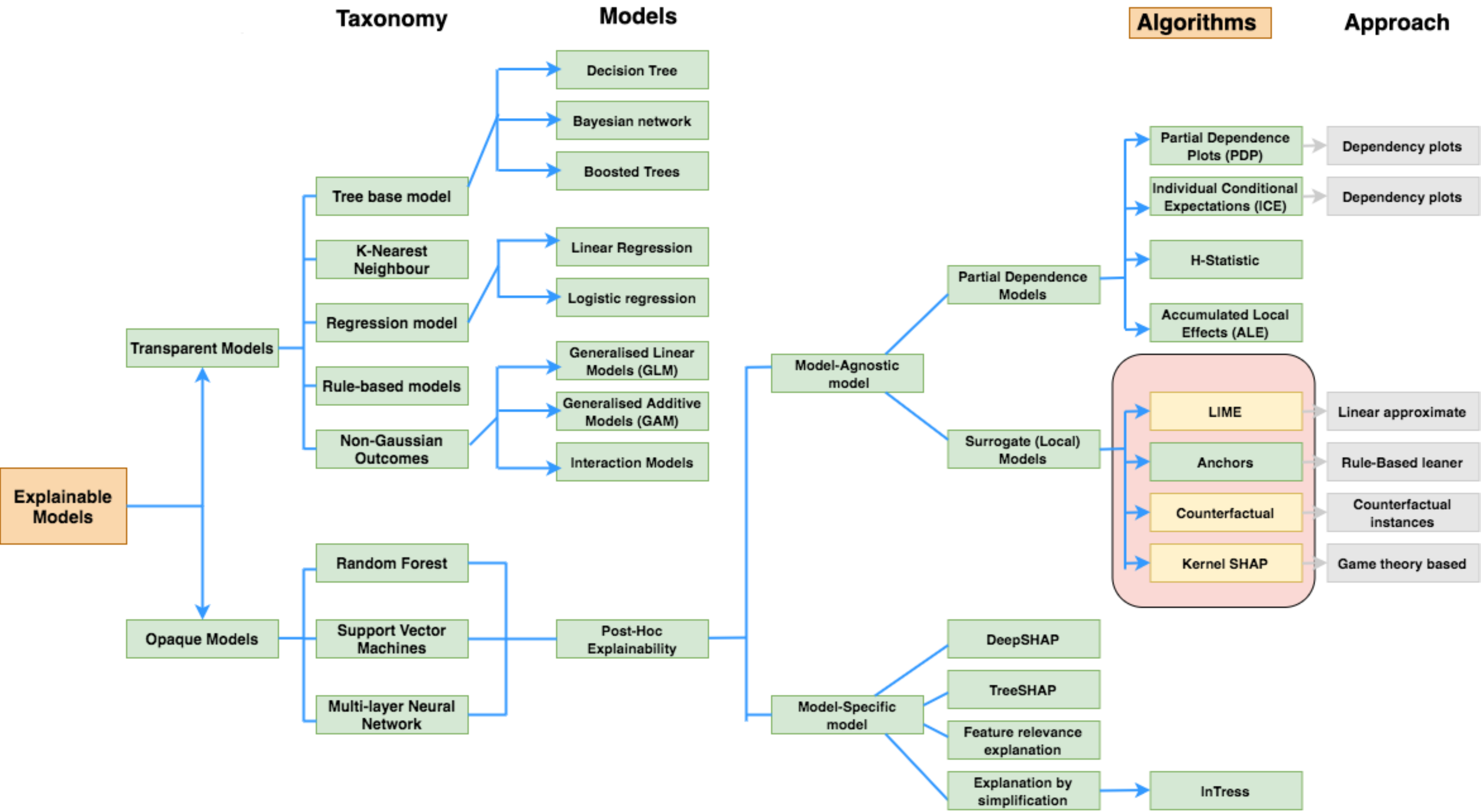}
    }
    \caption{\textcolor{blue}{Taxonomy of explainable artificial intelligence based on the taxonomy proposed by \citet{Belle20}}.}
    \label{fig:Taxonomy 2}
\end{figure}

Various approaches have been proposed in the literature to address the problem of interpretability in machine learning. Generally, this problem can be classified into two major models: interpretable models (inherently transparent) and model-agnostic, also referred to as post-hoc models (which aim to extract explanations out of opaque models). From our systematic literature review, these approaches can be categorised within the taxonomy presented in Figure~\ref{fig:Taxonomy 2}.

Interpretable models are by design already interpretable, providing the decision-maker with a transparent white box approach for prediction~\citep{Molnar18}. Decision trees, logistic regression, and linear regression are commonly used interpretable models. These models have been used to explain predictions of specific prediction problems~\citep{Siering2018}. Model-agnostic approaches, on the other hand, refer to the derivation of explanations from a black-box predictor by extracting information about the underlying mechanisms of the system \citep{Kim20dss}. %In addition, studies have focused on providing model-specific post-hoc explanations~. 

Model-agnostic models (post-hoc) are divided into two major approaches: partial dependency plots and surrogate models. The partial dependency plots can only provide pairwise interpretability by computing the marginal effect that one or two features have on the prediction. On the other hand, surrogate models consist of training a new local model that approximates the predictions of a black-box.
Model-agnostic post-hoc methods have the flexibility of being applied to any predictive model compared to model-specific post-hoc approaches. The two most widely cited post-hoc models in the literature include LIME~\citep{Ribeiro16} and Kernel SHAP~\citep{Lundberg17}. Counterfactuals can be generated using a post-hoc approach, and they can either be model-agnostic or model-specific. The main focus of this literature review is on model-agnostic post-hoc counterfactuals due to their flexibility and ability to work in any pre-existing trained model. This is detailed in Section~\ref{sec:counterfactuals_theory}.

\subsection{LIME - Local Interpretable Model-Agnostic Explanations}

Local Interpretable Model-agnostic Explanations (LIME)~\citep{Ribeiro16} explains the predictions of any classifier by approximating it with a locally faithful interpretable model. Hence, LIME generates local interpretations by perturbing a sample around the input vector within a neighborhood of a local decision boundary \citep{Elshawi19,Ribeiro16}. Each feature is associated with a weight computed using a similarity function that measures the distances between the original instance prediction and the predictions of the sampled points in the local decision boundary. An interpretable model, such as linear regression or a decision tree, can learn the local importance of each feature. This translates into a mathematical optimization problem expressed as

\begin{equation}
    explanation(x) = argmin_{g \in G}  \mathcal{L}(f, g, \pi_x) + \Omega(g),
    \label{eq:lime1}
\end{equation}

where $\mathcal{L}$ is the loss function which measures the similarity of the explainable model in the boundary of a perturbed data point $z$, $g(z)$, to the original black-box prediction, $f(z)$:

\begin{equation}
    \mathcal{L}(f, g, \pi_x) = \sum_{z,z' \in Z} \pi_x(z) \left( f(z) - g(z) \right)^2.
    \label{eq:lime2}
\end{equation}

In Equations~\ref{eq:lime1} and \ref{eq:lime2}, $x$ is the instance to be explained and $f$ corresponds to the original predictive black-box model (such as a neural network). $G$ is a set of interpretable models, where $g$ is an instance of that model (for instance, linear regression or a decision tree).  The proximity measure $\pi_x$ defines how large the neighborhood around instance x is that we consider for the explanation. Finally, $\Omega(g)$ corresponds to the model complexity, that is, the number of features to be taken into account for the explanation (controlled by the user)~\citep{Molnar18}.

\subsection{Approaches Based on LIME}

LIME has been extensively applied in the literature. For instance, \citet{Stiffler18} used LIME to generate salience maps of a certain region showing which parts of the image affect how the black-box model reaches a classification for a given test image~\citep{Zeiler13, Lapuschkin19}. \citet{Tan17} applied LIME to demonstrate the presence of three sources of uncertainty: randomness in the sampling procedure, variation with sampling proximity, and variation in the explained model across different data points. 

In terms of image data, the explanations are produced by creating a set of perturbed instances by dividing the input image into interpretable components (contiguous superpixels) and runs each perturbed instance via the model to get a probability~\citet{Malolan20}. After that, a simple linear model learns on this data set, which is locally weighted. At the end of the process, LIME presents the superpixels with the highest positive weights as an explanation. \citet{Preece18} proposed a CNN-based classifier, a LIME-based saliency map generator, and an R-CNN-based object detector to enable rapid prototyping and experimentation to integrate multiple classifications for interpretation.

Other researchers propose extensions to LIME. \citet{Turner16} derived a scoring system for searching the best explanation based on formal requirements using Monte Carlo algorithms. They considered that the explanations are simple logical statements, such as decision rules. \citep{Bastani17} utilized a surrogate model to extract a decision tree that represents the model behavior. \citep{Thiagarajan19} proposed an approach for building TreeView visualizations using a surrogate model. LIME has also been used to investigate the quality of predictive systems in predictive process analytics \citep{Sindhgatta20bpm}. In \citet{Sindhgatta20icsoc} the authors found that predictive process mining models suffered from different biases, including data leakage. They revealed that LIME could be used as a tool to debug black box models.

Lastly, a rule-based approach extension for LIME is Anchor \citep{Ribeiro18}. Anchor attempts to address some of the limitations by maximizing the likelihood of how a certain feature might contribute to a prediction. Anchor introduces IF-THEN rules as explanations and the notion of coverage, which allows the decision-maker to understand the boundaries in which the generated explanations are valid.

\subsection{SHAP - SHapley Additive exPlanations}

The SHAP (SHapley Additive exPlanations) is an explanation method that uses Shapley values \cite{Shapley52} from coalitional game theory to fairly distribute the gain among players, where contributions of players are unequal~\citep{Lundberg17}. Shapley values are a concept in economics and game theory and consist of a method to fairly distribute the payout of a game among a set of players. One can map these game-theoretic concepts directly to an XAI approach: a game is the prediction task for a single instance; the players are the feature values of the instance that collaborate to receive the gain. This gain consists of the difference between the Shapley value of the prediction and the average of the Shapley values of the predictions among the feature values of the instance to be explained \cite{Strumbelj13}. 

In SHAP, an explanation model, $g(z')$ is given by a linear combination of Shapley values $\phi_j$ of a feature $j$ with a coalitional vector, $z'_j$, of maximum size $M$,

\begin{equation}
g(z') = \phi_0 + \sum_{j=1}^{M} \phi_j z'_j.
\label{eq:shap1}
\end{equation}

\citet{Strumbelj13} claim that in a coalition game, it is usually assumed that $n$ players form a grand coalition that has a certain value. Given that we know how much each smaller (subset) coalition would have been worth, the goal is to distribute the value of the grand coalition among players fairly (that is, each player should receive a fair share, taking into account all sub-coalitions). \citet{Lundberg17} on the other hand, present an explanation using SHAP values and the differences between them to estimate the gains of each feature. 

To fairly distribute the payoff amongst players in a collaborative game, SHAP makes use of four fairness properties: (1) Additivity, which states that amounts must sum up to the final game result, (2) Symmetry, which states that if one player contributes more to the game, (s)he cannot get less reward, (3) Efficiency, which states that the prediction must be fairly attributed to the feature values, and (4) Dummy, which says that a feature that does not contribute to the outcome should have a Shapley value of zero.

\subsection{Approaches Based on SHAP}

In terms of related literature, \citet{Ariza20} adopted SHAP values to assess the logistic regression model and several machine learning algorithms for granting scores in P2P (peer-to-peer) lending; the authors point out SHAP values can reflect dispersion, nonlinearity and structural breaks in the relationships between each feature and the target variable. They concluded that SHAP could provide accurate and transparent results on the credit scoring model. \citet{Parsa20} also highlight that SHAP could bring insightful meanings to interpret prediction outcomes. For instance, one of the techniques in the model, XGBoost, not only is capable of evaluating the global importance of the impacts of features on the output of a model, but it can also extract complex and non-linear joint impacts of local features.

Recently, \citet{Wang2021shapley} proposed to generalise the notion of Shapley value axioms for directed acyclic graphs. The new algorithm is called Shapley flow and relies on causal graphs in order to be able to compute the flow of Shapley values that describe the internal mechanisms of the black-box.

In the following sections, we will expand the analysis of model-agnostic approaches for XAI by conducting a systematic literature review on counterfactual and causability approaches for XAI.

\section{Systematic Literature Review Towards Counterfactuals and Causability in XAI}\label{sec:systematic_review}

The purpose of this systematic review paper is to investigate the theories, algorithms, and applications that underpin XAI approaches that have the potential to achieve \textit{causability}. This means that this paper will survey the approaches in the extensive body of literature that are primarily based on causality and counterfactuals to help researchers identify knowledge gaps in the area of interest by extracting and analysing the existing approaches. 

Our systematic literature review follows the Preferred Reporting Items for Systematic Reviews and Meta-Analyses (PRISMA) framework as a standardized way of extracting and synthesizing information from existing studies concerning a set of research questions. More specifically, we followed the PRISMA checklist\footnote{\url{http://www.prisma-statement.org/documents/PRISMA\%202009\%20checklist.pdf}} with the study search process presented in the PRISMA flow diagram\footnote{\url{http://prisma-statement.org/documents/PRISMA\%202009\%20flow\%20diagram.pdf}}. 

Based on PRISMA, the procedure of systematic review can be separated into several steps: (1) definition of the research questions; (2) description of the literature search process and strategy. Inspired in the recent work of~\citet{Teh2020}, we conducted a topic modeling analysis to refine the search results using the Latent Dirichlet Allocation (LDA) algorithm together with an inclusion and exclusion criteria to assist with the selection of relevant literature; (3) extraction of publication data (title, abstract, author keywords and year), systemisation, and analysis of the relevant literature on counterfactuals and causality in XAI; (4) Lastly, we conducted identification of biases and limitations in our review process.

\subsection{Research Questions}

To help researchers identify knowledge gaps in the area of causality, causability, and counterfactuals in XAI, we proposed the following research questions:

\begin{itemize}
\item RQ1: What are the main theoretical approaches for counterfactuals in XAI (Theory)?

\item RQ2: What are the main algorithms in XAI that use counterfactuals as a means to promote understandable causal explanations (Algorithms)?

\item RQ3: What are the sufficient and necessary conditions for a system to promote causability (Applications)?

\item RQ4: What are the pressing challenges and research opportunities in XAI systems that promote Causability?

\end{itemize}

\subsection{Search Process}

To address the proposed research questions, in this paper, we used three well-known Computer Science academic databases: (1) Scopus, (2) IEEE Xplore, and (3) Web of Science (WoS). We considered these databases because they have good coverage of works on artificial intelligence, and they provide APIs to retrieve the required data with few restrictions. We used the following search query to retrieve academic papers in artificial intelligence related to explainability or interpretability and causality or counterfactuals.

\begin{center}
\vspace{0.1cm}
( \textit{artificial} AND \textit{intelligence} ) AND ( \textit{xai} OR \textit{explai*} OR \textit{interpretab}* ) AND ( \textit{caus}* OR \textit{counterf}* )
\vspace{0.1cm}
~\\*
\end{center}

\begin{figure}[h!]
    \centering
    \includegraphics[scale=0.5]{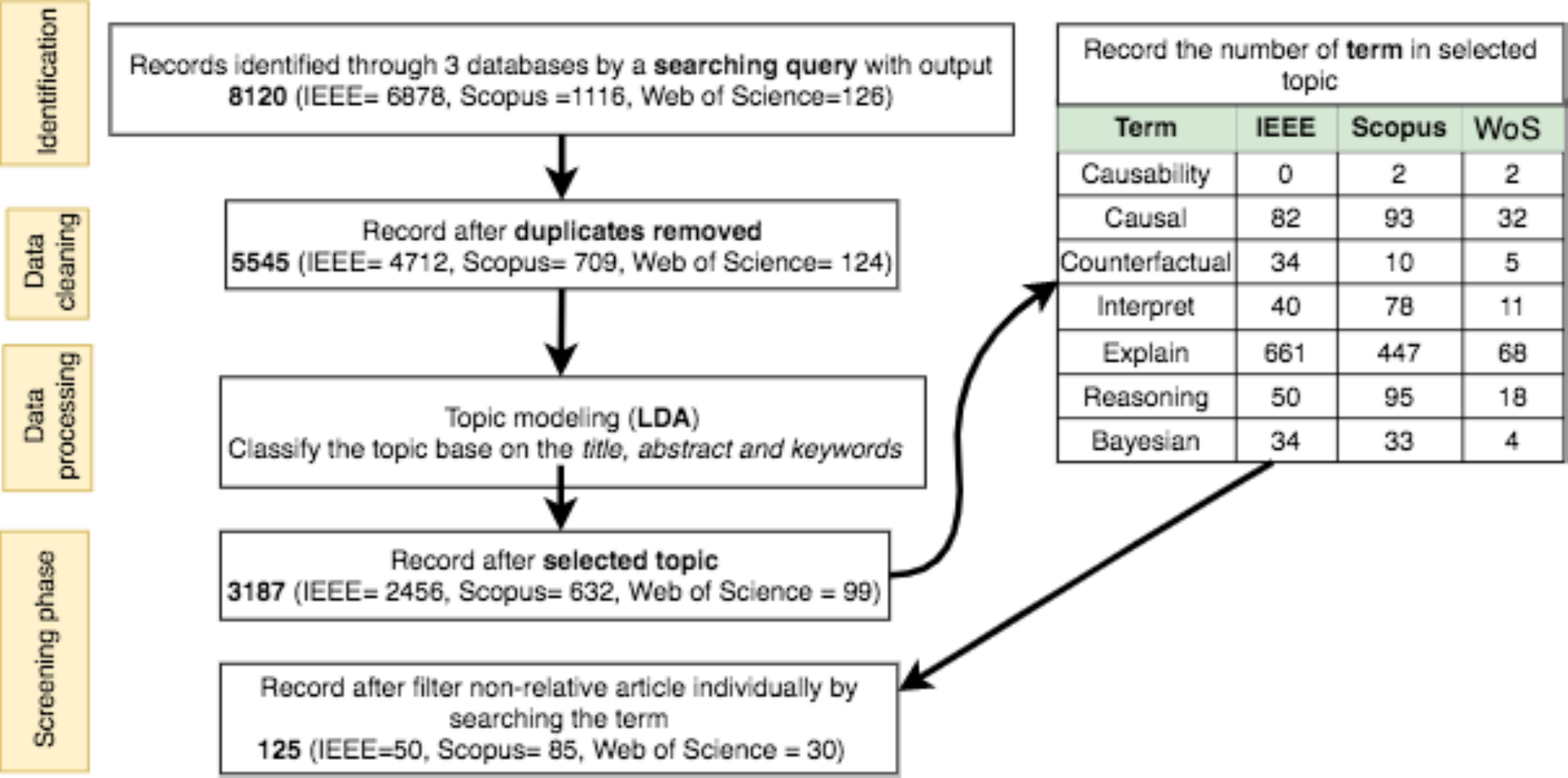}
    \caption{PRISMA flow diagram search results.}
    \label{fig:prisma_summary}
\end{figure}

This query allowed us to extract bibliometric information from different databases, such as publication titles, abstracts, keywords, year, etc. The initial search returned the following articles: IEEE Xplore (6878), Scopus (116), WoS (126). We removed duplicate entries in these results as well as results that had missing entries. In the end, we reduced our search process to IEEE Xplore (4712), Scopus (709), WoS (124). Our strategy is summarised in the PRISMA flow diagram illustrated in Figure~\ref{fig:prisma_summary}.

To guarantee that the initial query retrieved publications that match this review's scope, we conducted a topic modelling analysis based on Latent Dirichlet Allocation (LDA) to refine our search results. 

\subsection{Topic Modelling}

Topic modelling is a natural language processing technique that consists of uncovering a document collection's underlying semantic structure based on a hierarchical Bayesian analysis. LDA is an example of a topic model used to classify text in a document to a particular topic. It builds a topic per document model and words per topic model, modelled as Dirichlet distributions. In our search strategy, LDA enabled us to cluster words in publications with a high likelihood of term co-occurrence and allowed us to interpret the topics in each cluster. This process guaranteed that the papers classified within a topic contain all the relevant keywords to address our research questions.  

\begin{figure}[!h]
   \resizebox{\columnwidth}{!} {

    \includegraphics{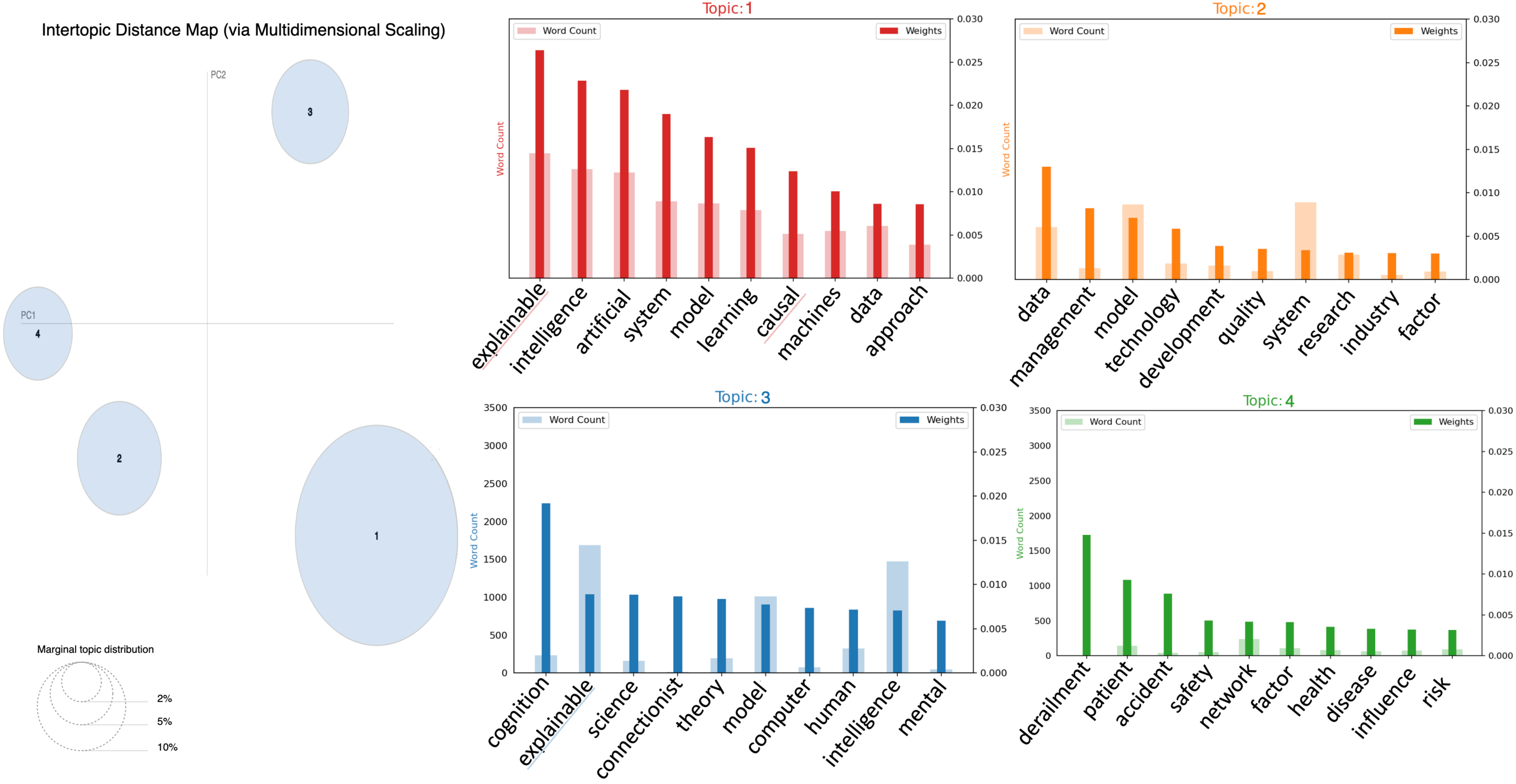}
    }
    \caption{Best performing LDA topic model for Scopus database, using 709 titles, abstracts, and authors keywords found from the proposed search query. Figure also shows the top 10 most relevant words for each Topic. }
    \label{fig:lda}
\end{figure}

In this paper, we used the title, abstract, and authors' keywords retrieved from the proposed query, and applied several text mining techniques, such as stop word removal, word tokenisation,  stemming, and lemmatisation.  We then analysed the term co-occurrences with LDA for each database. The best-performing model contained a total of 4 topics. The LDA model's output is illustrated in Figure~\ref{fig:lda} with the inter-topic distance showing the marginal topic distributions (left) and the top 10 most relevant terms for each topic. Analysing Figure~\ref{fig:lda}, Topic 1 contained all the words that are of interest to the research questions proposed in this survey paper: explainability, causality, and artificial intelligence. Topic 2, on the other hand, has captured words that are primarily related to data management and technology. Topic 3 has words related to the human aspect of explainable AI, such as cognition, mental, and human. Finally, Topic 4 contains words associated with XAI in healthcare. For this survey paper, we chose all the publications classified as either Topic 1 or Topic 3. In the end, we were able to reduce our search results to IEEE Xplore (3187), Scopus (632), WoS (99). After manually looking at these publication records and selecting articles about "causability", "causal", "counterfactual", we obtained our final set of documents for analysis: IEEE Xplore (125), Scopus (85), WoS (30).

\subsection{Word Co-Occurrence Analysis}

In our survey, we are interested in understanding the necessary and sufficient conditions to achieve causability and how current approaches can promote it. We started to analyse the keyword co-occurrence in the returned documents from our search query to achieve this understanding. We collected the title, abstract, and authors' keywords from the search results in $Scopus$, and filtered the results from using three different keywords of interest: explainable AI, counterfactuals, and causality. This resulted in three different $Scopus$ files with the keywords of interest. 

To visualise the results, we used the graphical capabilities of \textit{VOS Viewer}\footnote{https://www.vosviewer.com/}, which is a software tool for constructing and visualising bibliometric networks. 

\begin{figure}[!h]
    \centering
     \resizebox{\columnwidth}{!} {
    \includegraphics{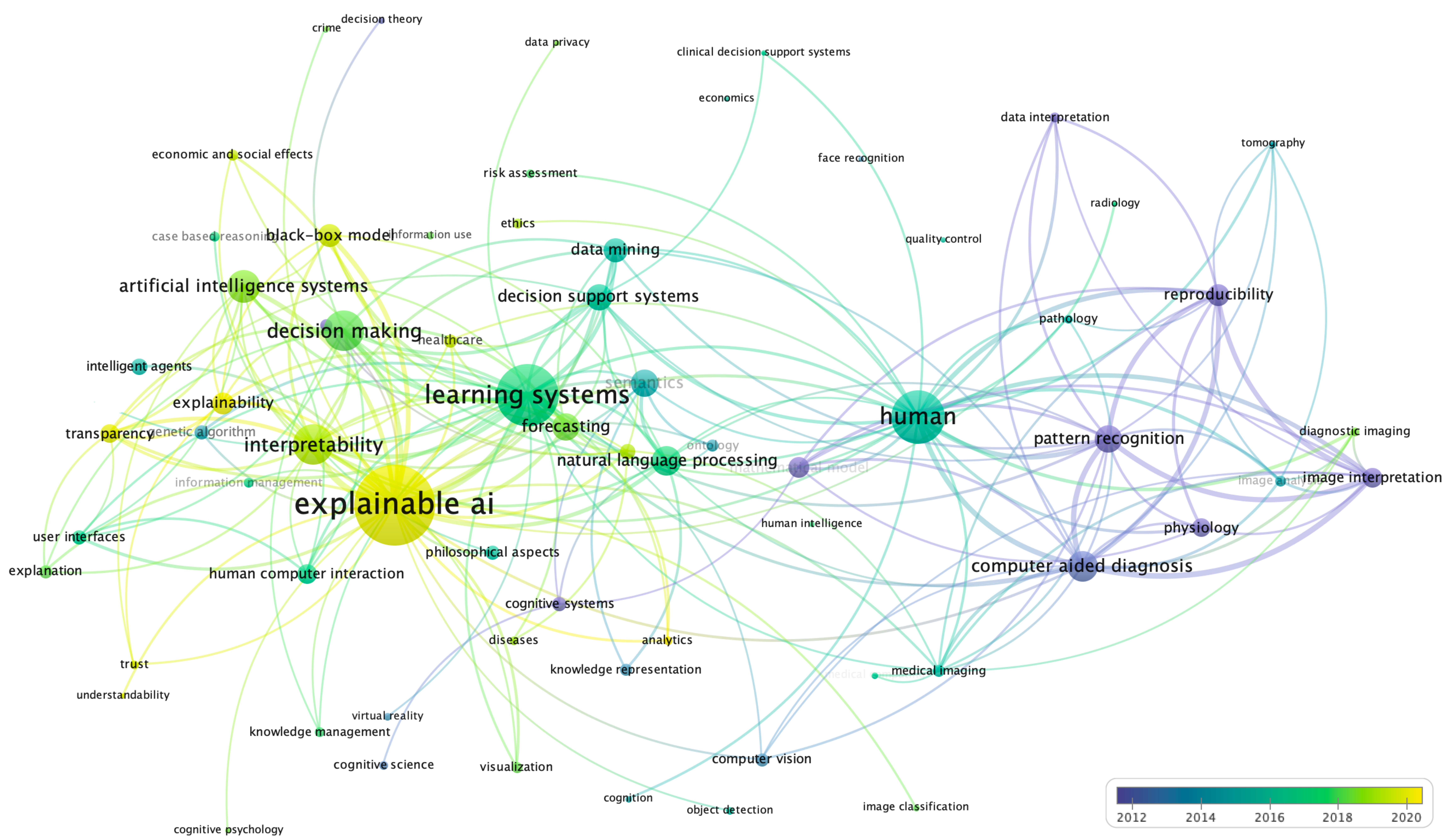}
    }
    \caption{Network visualization of co-occurrence between keywords in articles about XAI.}
    \label{fig:vos_xai}
\end{figure}

Figure~\ref{fig:vos_xai} represents the co-occurrence of authors' keywords regarding the field of XAI. The density plot reveals a shift in research paradigms evolving from \textit{machine-centric} topics to more \textit{human-centric} approaches involving intelligent systems and cognitive systems, to the need of \textit{explainability} in autonomous decision-making. 

It is interesting to note that machine-centric research interests (such as pattern recognition or computer-aided diagnostic systems) started to change around 2016. The European Union Commission started to put forward a long list of regulations for handling consumer data, the GDPR. In that year, publications start shifting their focus from fully autonomous systems to a human-centric view of learning systems with a need for interpretability in decision-making. Figure~\ref{fig:vos_xai} also shows another shift of research paradigms around 2018 towards \textit{explainable AI}, which coincides with the year where GDPR was put into effect in the European Union, imposing new privacy and security standards regarding data access and usage.  One of these standards is Article 22, which states that an individual has "the right not to be subject to a decision based solely on automated processing"\footnote{\url{https://www.privacy-regulation.eu/en/article-22-automated-individual-decision-making-including-profiling-GDPR.htm}}. In other words, an individual has the right for explainability whenever a decision is computed from an autonomous intelligent system. Given that these systems are highly opaque with complex internal mechanisms, there has been a recent growing need for \textit{transparent} and \textit{interpretable} systems that are able to secure \textit{ethics} and promote user \textit{understandability} and \textit{trust}.

\begin{figure}[!h]
    \centering
    \resizebox{\columnwidth}{!} {
    \includegraphics{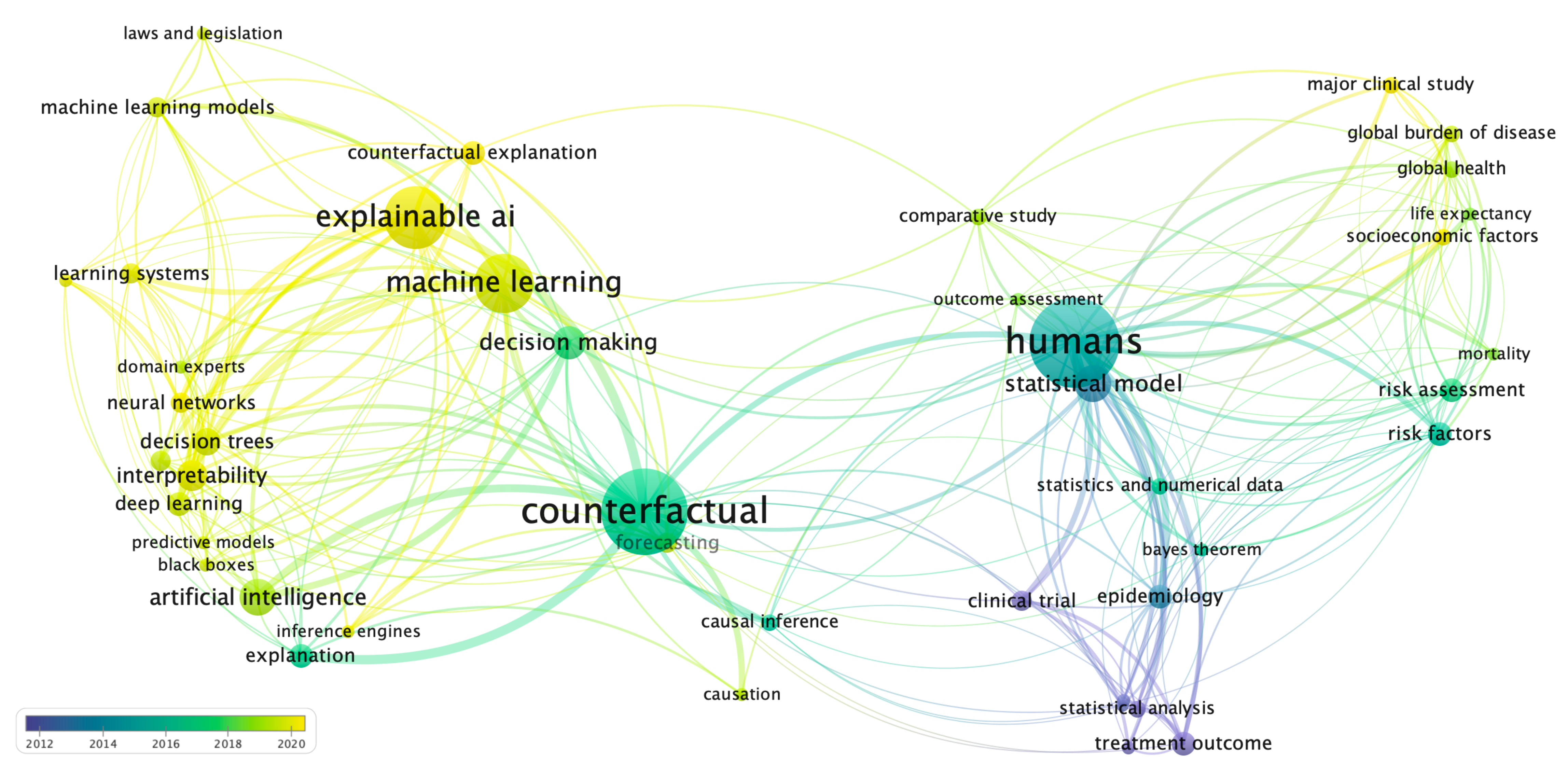}
    }
    \caption{Network visualization of co-occurrence between keywords in articles about counterfactuals in XAI. }
    \label{fig:vos_counterfactual}
\end{figure}

Some researchers argued that for a machine to achieve a certain degree of human intelligence, and consequently, explainability, then counterfactuals need to be considered~\citep{Holzinger19,Holzinger20}. Recently, \citet{Miller2019} stated that explanations need to be counterfactuals (‘‘contrary-to-fact’’)~\citep{Byrne1997}, since they enable mental representations of an event that happened and also representations of some other event alternative to it~\citep{Stepin2021}. Counterfactuals describe events or states of the world that did not occur and implicitly or explicitly contradict factual world knowledge. For instance, in cognitive science, counterfactual reasoning is a crucial tool for children to learn about the world~\citep{Weisberg13}. The process of imagining a hypothetical scenario of an event that is contrary to an event that happened and reasoning about its consequences is defined as \textit{counterfactual reasoning}~\citep{Pereira2020}. 
%Counterfactual reasoning is key for explaining adaptive behaviour in a changing environment~\citep{Paik2014}. 
We investigated the word co-occurrence in articles involving \textit{explainable AI} and \textit{counterfactuals} to understand how the literature is progressing in this area. Figure~\ref{fig:vos_counterfactual} shows the obtained results.

\begin{figure}[!h]
    \centering
    \resizebox{\columnwidth}{!} {
    \includegraphics{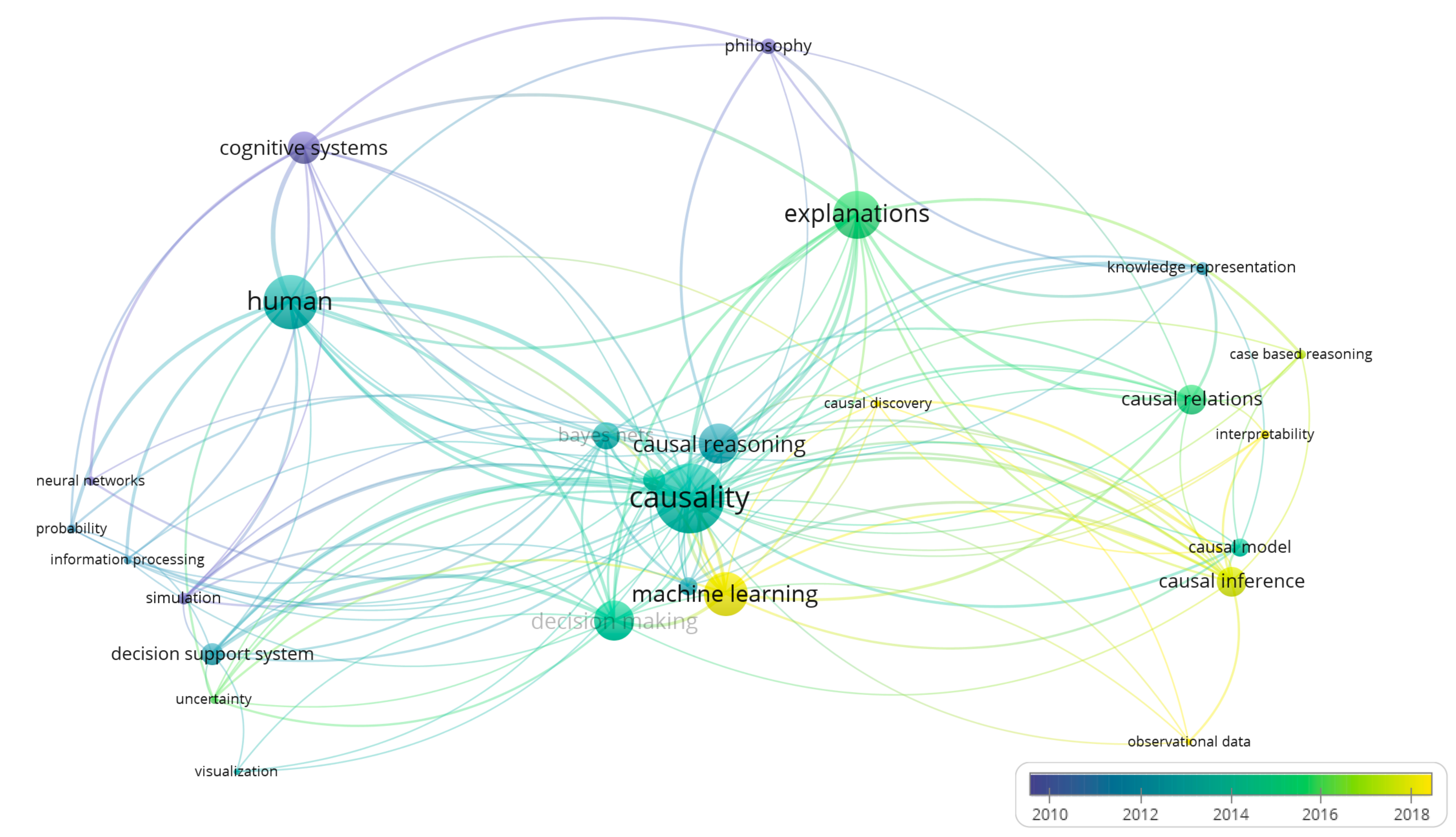}
    }
    \caption{Network visualization of co-occurrence between keywords in articles about causality in XAI. }
    \label{fig:vos_causality}
\end{figure}

In the density plot in Figure~\ref{fig:vos_counterfactual}, one can see that counterfactual research in XAI is a topic that has gained interest in the scientific community very recently, with most of the scientific papers dating from 2019 on-wards. This reflects the need for supporting explanations with contrastive effects: by asking ourselves what would have been the effect of something if we had not taken action, or vice versa. Creating such hypothetical worlds may increase the user's understanding of how a system works. The figure seems to be suggesting that the recent body of literature concerned with counterfactuals for XAI is motivated by the medical decision-making domain since we can see relevant keywords such as \textit{patient treatment}, \textit{domain experts}, and \textit{diagnosis}. There is also a recent body of literature in clinical research supporting the usage of counterfactuals and causality to provide interpretations and understandings for predictive systems~\citep{Prosperi2020}. 

Some researchers also argued that for a machine to achieve a certain degree of human intelligence, causality must be incorporated in the system~\citep{Pear19}. Others support this idea in the context of XAI, where they argue that one can only achieve a certain degree of explainability if there is a causal understanding of the explanations, in order words, if the system promotes causability~\citep{Holzinger19}. In this sense, we also analysed co-occurrence between keywords in articles about causality in XAI. Figure~\ref{fig:vos_causality} illustrates the results obtained.

In terms of causality, Figure~\ref{fig:vos_causality}, one can draw similar conclusions. Although the figure shows a clear connection between Artificial Intelligence and causality (causal reasoning, causal graphs, causal relations), the literature connecting causal relations to explainable AI is scarce. This opens new research opportunities in the area, where we can see from  Figure~\ref{fig:vos_causality} a growing need for counterfactual research. Literature regarding $causability$ seems to be also very scarce and very recent. New approaches are needed in this direction, and it is the purpose of this systematic review to understand which approaches for XAI are underpinned by causal theories.

\subsection{Inclusion and exclusion criteria}

To select relevant literature from the obtained search results, we had to consider which papers should be included in our analysis and which ones should be excluded in order to be able to address the proposed research questions. Table~\ref{tab:criteria} summarises the selected criteria.

\begin{table}[!h]
\centering
\begin{tabular}{|l|l|}
\hline
\textbf{Inclusion Criteria}         & \textbf{Exclusion Criteria}          \\ \hline
Papers about causality in XAI       & Papers about causal machine learning \\ \hline
Papers about counterfactuals in XAI & Papers about causality               \\ \hline
Papers about causability            & Papers not in English                \\ \hline
Papers about main algorithms in XAI & Papers without algorithms for XAI    \\ \hline
\end{tabular}
\caption{Inclusion and exclusion criteria to assess the eligibility of research papers to analyse in our systematic literature review.}
\label{tab:criteria}
\end{table}

\subsection{Risk of bias}

As with any human-driven task, the process of finding relevant research is affected by cognitive biases. In this systematic review, we acknowledge that limiting our search to three databases (Scopes, Web of Science, and IEEE) might have contributed to missing articles. Databases that could have complemented our search could be Google Scholar, SpringerLink, and PubMed. Another consideration is that we did not extract the references from the collected papers to enrich our search. The collection of retrieved documents was already too big, and we found that doing this would exponentially increase the complexity of the LDA topic analysis that we conducted. Finally, the search query was restricted to keywords relevant to collecting the papers of interest. These keywords, however, might have limited our search, and we might have missed relevant articles.

\section{Counterfactual Approaches Explainable AI: Properties for Good Counterfactuals}\label{sec:properties}

The systematic review that we conducted allowed us to understand the different counterfactual approaches for XAI. As mentioned throughout this article, counterfactuals have been widely studied in different domains, especially in philosophy, statistics, and cognitive science. Indeed, researchers are arguing that counterfactuals are a crucial missing component that has the potential to provide a certain degree of human intelligence and human-understandable explanations to the field of XAI~\citep{Holzinger19}. Other researchers state that counterfactuals are essential to elaborate predictions at the instance-level~\citep{Sokol20} and to make decisions actionable~\citep{Fernandez19}. Other researchers claim that counterfactuals can satisfy GDPR's legal requirements for explainability~\citep{Wachter18}.

\begin{figure}[!h]
    \centering
    \resizebox{0.7\columnwidth}{!} {
    \includegraphics{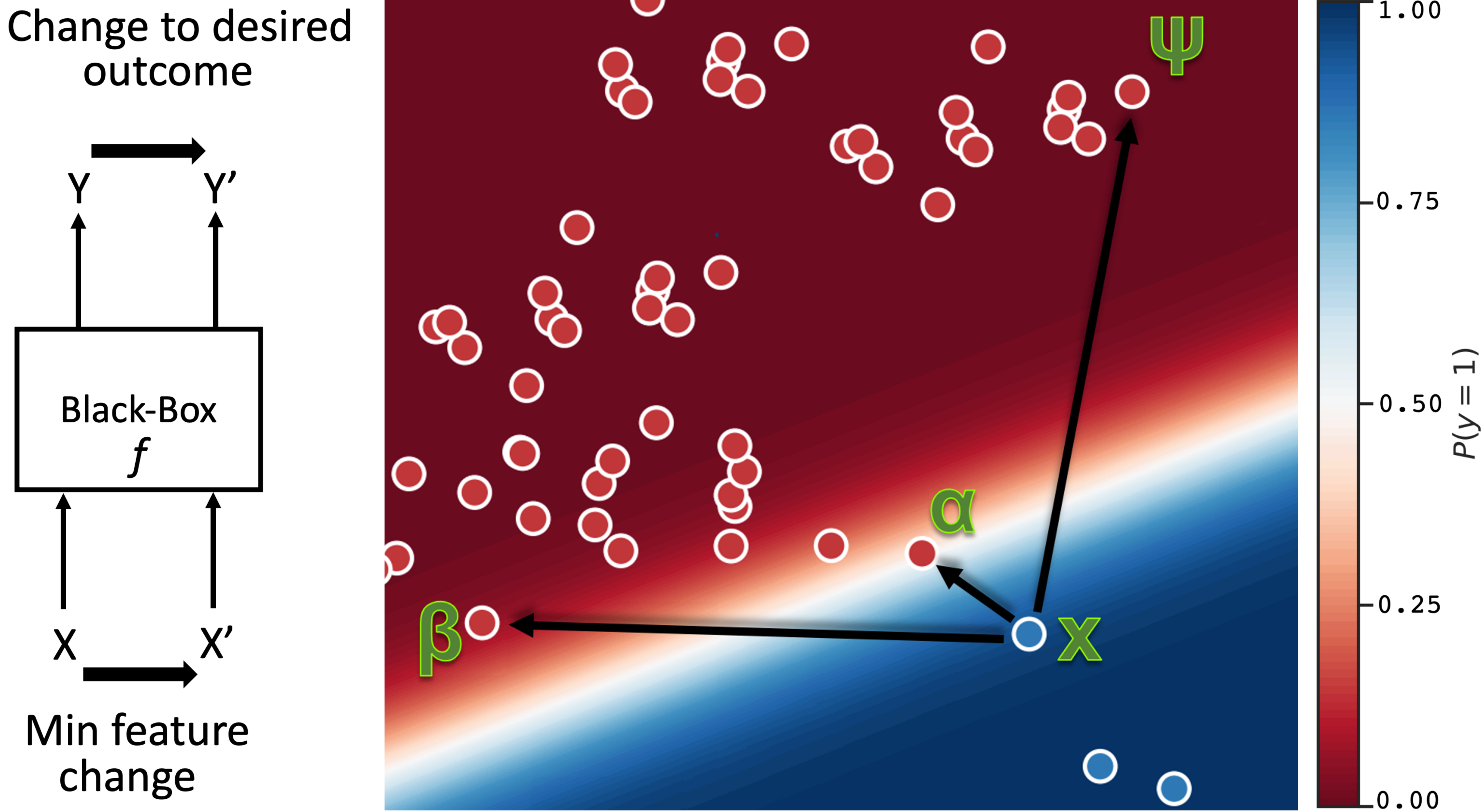}
    }
    \caption{Different counterfactual candidates for data instance $x$. According to many researchers, counterfactual $\alpha$ is the best candidate, because it has the smallest Euclidean distance to $x$~\citep{Wachter18}. Other researchers argue that counterfactual instance $\gamma$ is the best choice since it provides a feasible path from $x$ to $\gamma$~\citep{Poyiadzi20}. Counterfactual $\beta$ is another candidate of poor quality because it rests in a less defined region of the decision boundary.}
    \label{fig:counterfactual_general}
\end{figure}

Most XAI algorithms attempt to achieve explainability by (1) perturbing a single data instance, generating a set of perturbed data points around the decision boundary, (2) passing these perturbed instances through the black box, generating labels to these data points, and by (3) fitting an interpretable model (such as linear regression or a decision tree) to the perturbed data points~\citep{Ribeiro16}. Counterfactuals are classified as \textit{example-based approaches} for XAI~\citep{Molnar18}. They are based on approaches that compute which changes should be made to the instance datapoint to flip its prediction to the desired outcome \citep{Wachter18}. 
Figure~\ref{fig:counterfactual_general} shows an illustration of several counterfactual candidates for a data instance $x$ according to different works in the literature~\citep{Poyiadzi20}.

\subsection{The Importance of Distance Functions in Counterfactual Approaches for XAI}\label{sec:distances}

The definition of counterfactual as the minimum distance (or change) between a data instance and a counterfactual instance goes back to the theory proposed by  \citet{Lewis73counterfactual}. Given a data point $x$, the closest counterfactual $x'$ can be found by solving the problem where $d(.,.)$ is a measurement for calculating the distance from the initial point to the generated point.

\begin{equation}
 argmin_{x'} d (x,x')
\label{eq:initialCF}
\end{equation}   

One important question that derives from Equation~\ref{eq:initialCF} is \textit{what kind of distance function should be used?} Different works in the literature address this optimization problem by exploring different distance functions and $L_p$-norms. This section will review different norms used as distance functions in the literature of XAI and their properties.

In general, a norm measures the size of a vector, but it can also give rise to distance functions. The $L_p$-norm of a vector $x$ is defined as:
\begin{equation}
\left| \left| x \right|\right|_p = \left( \sum_{i=1}^n \left| x_i \right|.^p  \right)^{1/p}
\label{eq:norms}
\end{equation}

\begin{figure}[!h]
    \centering
    \resizebox{0.7\columnwidth}{!} {
    \includegraphics{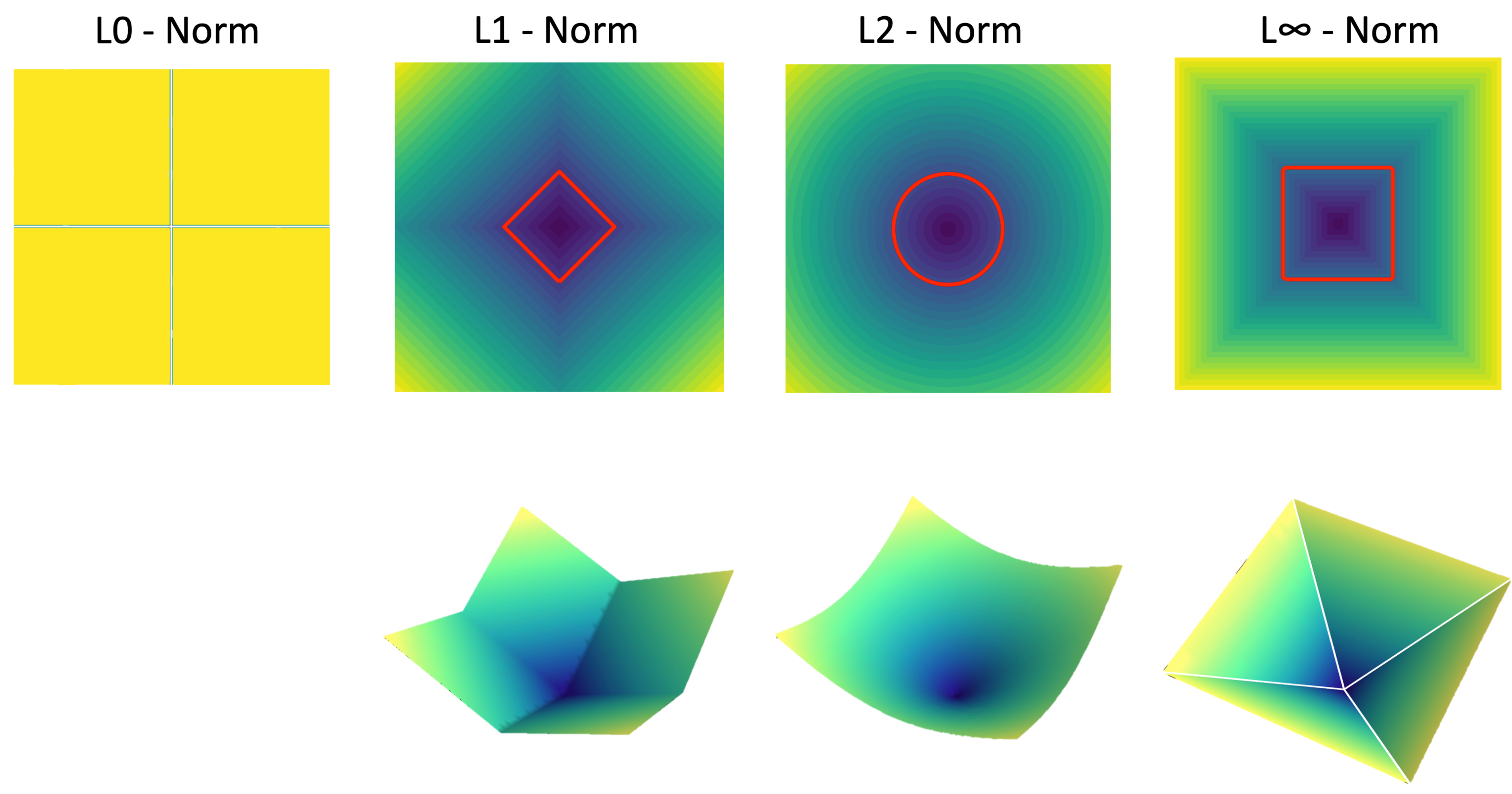} 
    }
    \caption{Graphical visualisation of different $L_p$-norms: $L_0$-norms (which is not really a norm by definition), the $L_1$-norm (also known as Manhattan distance), the $L_2$-norm (known as the Euclidean distance, and the $L_\infty$-norm.}
    \label{fig:lnorms}
\end{figure}

Equation~\ref{eq:norms} shows that different values of $p$ yields a different distance function with specific properties. The systematic literature review revealed that most works in XAI used either the $L_0$-norm (which is not a norm by definition), the $L_1$-norm (also known as Manhattan distance), the $L_2$-norm (known as the Euclidean distance, and the $L_\infty$-norm. Figure~\ref{fig:lnorms} shows a graphical representation of the different norms and the respective contours. 

\begin{itemize}
    \item \textbf{$L_0$-norm}. The $L_0$-norm has been explored in the context of counterfactuals in XAI primarily by \citet{Dandl2020} and \citet{Karimi2020mace}. Given a vector $x$, it is defined as
    \begin{equation}
        \left| \left| x \right| \right|_0 = \sqrt[0]{ \sum_i x_i^0}.
    \end{equation}
    Intuitively, $L_0$-norm is the number of nonzero elements in a vector, and it is used to count the number of features that change between the initial instance $x$ and the counterfactual candidate $x'$, resulting in sparse counterfactual candidates~\citep{Karimi2020mace}. Figure~\ref{fig:lnorms} shows a visualisation of the $L_0$-norm where one can see that the function is completely undifferentiable, making it very hard to find efficient solutions to minimize it. 
    
    \item \textbf{$L_1$-norm}. The $L_1$-norm (also known as the Manhattan distance) has been the most explored distance function in the literature of counterfactuals in XAI. \citet{Wachter18} argued that the $L_1$-norm provides the best results for finding good counterfactuals, since it induces sparse solutions. Given a vector $x$, the $L_1$-norm is defined as
    \begin{equation}
        \left| \left| x \right| \right|_1 = \sum_i \left| x_i \right|.
    \end{equation}
    Intuitively, $L_1$-norm is used to restrict the average change in the distance between the initial instance $x$ and the counterfactual candidate $x'$. Since the $L_1$-norm gives an equal penalty to all parameters and leads to solutions with more large residuals, it enforces sparsity. In Figure~\ref{fig:lnorms}, one can see that the major problem with $L_1$-norms is its diamond shape, which makes it hard to differentiate.

    \item \textbf{$L_2$-norm}. The $L_2$-norm (also known as the Euclidean distance) has been one of the most explored distance functions in the literature of counterfactuals in XAI, although it does not provide sparse solutions when compared with the $L_1$ or $L_0$-norm. Given a vector $x$, the $L_1$-norm is defined as
    \begin{equation}
        \left| \left| x \right| \right|_2 = \sqrt{\sum_i x_i^2}.
    \end{equation}
    Intuitively, the $L_2$-norm measures the shortest distance between two points and can detect a much larger error than the $L_1$-norm, making it more sensitive to outliers. Although the $L_2$-norm does not lead to sparse vectors, it has the advantage that it is differentiable. Figure~\ref{fig:lnorms} shows that smoothness and rotational invariance (a circle or a hyper-sphere in higher dimensions) are both desirable properties in many optimization problems, making it computationally efficient. %The $L_2$-norm can be seen as a norm that is in the middle of the strict diamond shape $L_1$-norm and the $L_\infty$-norm.
    
    \item \textbf{$L_\infty$-norm}. The $L_\infty$-norm has been explored in the context of counterfactuals in XAI primarily by \citet{Karimi2020mace}. Given a vector $x$, it is defined as
    \begin{equation}
        \left| \left| x \right| \right|_\infty = \sqrt[\infty]{\sum_i x_i^\infty} = max(\left| x_i \right| ).
    \end{equation}
    Intuitively, $L_\infty$-norm is used to restrict maximum change across features. The maximum change across the features between the initial instance $x$ and the counterfactual candidate $x'$~\citep{Karimi2020mace}. Computationally, the $L_\infty$-norm is differentiable in every point, except when at least two features $x_i$ have the same absolute values $\left| x_i \right|$, which is illustrated in Figure~\ref{fig:lnorms}. By minimizing the L-infinity norm, we are penalizing the cost of the largest feature, leading to less sparse solutions compared to $L_0$-norm or $L_1$-norm.

\end{itemize}

Distance functions in counterfactuals are associated with the sparsity of the vector, which is a highly desirable property to have when looking for counterfactuals. The minimum the changes we can have in the features, the better and more human interpretable counterfactuals we will find. The following section will present the main properties that a theory for counterfactuals in XAI should satisfy. 

\subsection{Properties to Generate Good Counterfactuals}

\textcolor{blue}{Literature suggests a set of properties that need to be satisfied in order to generate a good (interpretable) counterfactual:}

\begin{itemize}
     \item \textbf{Proximity.}
     Proximity calculates the distance of a counterfactual from the input data point while generating a counterfactual explanation, \citep{Verma20}. As mentioned in Section~\ref{sec:distances}, many different distance functions can be used to measure proximity, resulting in counterfactual candidates with different properties. Other works in the literature consider another types of proximity measures such as Nearest Neighbour Search~\citep{Keane20}, cosine similarity~\citep{Martens14}, or even learning to rank techniques \citep{Moreira11,Moreira11Inforum}.

     \item \textbf{Plausibility.} 
     This property is similar to the terms \textit{Actionability} and \textit{Reasonability} referred in  \cite{Keane2020, Verma20}. It emphasizes that the generated counterfactuals should be legitimate, and the search process should ensure logically reasonable results. This means that a desirable counterfactual should never change \textit{immutable} features such as gender or race. When explaining a counterfactual, one cannot have explanations like \textit{"if you were a man, then you would be granted a loan"}, since these would show an inherent bias in the explanation. Mutable features, such as income, should be changed instead to find good counterfactuals. 
    
\item  \textbf{Sparsity.}
     This property is related to the methods used to efficiently find the minimum features that need to be changed to obtain a counterfactual~\citep{Keane2020}. 

    In cognitive science, counterfactuals are used as a process of imagining a hypothetical scenario contrary to an event that happened and reasoning about its consequences \citep{Pereira2020}. 
    \textcolor{blue}{It is desired that counterfactuals are sparse, i.e., with the fewest possible changes in their features. This property leads to more effective, human-understandable, and interpretable counterfactuals.} In \citet{Mothilal20}, for instance, the authors elaborate that sparsity is assessing how many features a user needs to change to transition to the counterfactual class. On the other hand, \citet{Verma20} argues that sparsity can be seen as a  trade-off between the number of features and the total amount of change made to obtain the counterfactual. \citet{Wachter18} also stands on this idea and asserts that pursuing the "closest possible world", or the smallest (minimum-sized) change to the world that can be made to obtain a desirable outcome. 
    
    \textcolor{blue}{Recently, \citet{Pawelczyk2020multiplicity} proposed a theoretical framework that challenges the notion that counterfactual recommendations should be sparse. The authors argue the problem of predictive multiplicity can result in situations where there does not exist one superior solution to a prediction problem with respect to a measure of interest (e.g. error rate).} 
     
\item  \textbf{Diversity.} 
This property was introduced in the work of~\citet{Russell19} and also explored in~\citet{Mothilal20,Karimi2021survey}. Finding the closest points of an instance $x$ according to a distance function can lead to very similar counterfactual candidates with small differences between them. Diversity was introduced as the process of generating a set of diverse counterfactual explanations for the same data instance $x$~\citep{Karimi2020mace}. This leads to explanations that are more interpretable and more understandable to the user.

\item \textbf{Feasibility.} 
This property was introduced by \citet{Poyiadzi20} as an answer to the argument that finding the closest counterfactual to a data instance does not necessarily lead to a feasible change in the features. In Figure~\ref{fig:counterfactual_general}, one can see different counterfactual candidates. The closest counterfactual to the data instance $x$ is $\alpha$. However, this point falls in the decision boundary. Thus, the black-box is not very certain about its class, which could lead to biased counterfactual explanations. 
To address this problem, \citet{Poyiadzi20} argues that counterfactual $\gamma$ is a better one because it falls in a well-defined region of the decision boundary and also corresponds to the point that has the shortest path to $x$. This way, it is possible to generate human-interpretable counterfactuals with the least possible feature changes.
\end{itemize}

\textcolor{blue}{From the definitions of plausibility and feasibility, one can conclude that they are related to each other: for a counterfactual to be feasible, it needs first to be plausible.  Plausibility refers to a property that ensures that generated counterfactuals are legitimate. This means that a legitimate counterfactual should never change immutable features such as gender or race. On the other hand, feasibility is related to the search of a counterfactual that does not lead to ''paradoxical interpretation''. Using the example from \citet{Poyiadzi20}, low-skilled unsuccessful mortgage applicants may be told to double their salary. This implies that they need to increase their skill level first, which may lead to counterfactual explanations that are impractical and, therefore, infeasible. Thus, satisfying feasibility automatically guarantees plausible counterfactuals, promoting a higher level of interpretability of counterfactual explanations.}

Given the above properties, in the following sections, we will classify the different algorithms found in the literature by (1) their underlying theory (Section~\ref{sec:counterfactuals_theory}), and by (2) the above properties (Section~\ref{sec:counterfactuals_algo}).

\section{Counterfactual Approaches in Explainable AI: The Theory}\label{sec:counterfactuals_theory}

\begin{figure}[!h]
\centering
    \resizebox{\columnwidth}{!} {
    \includegraphics{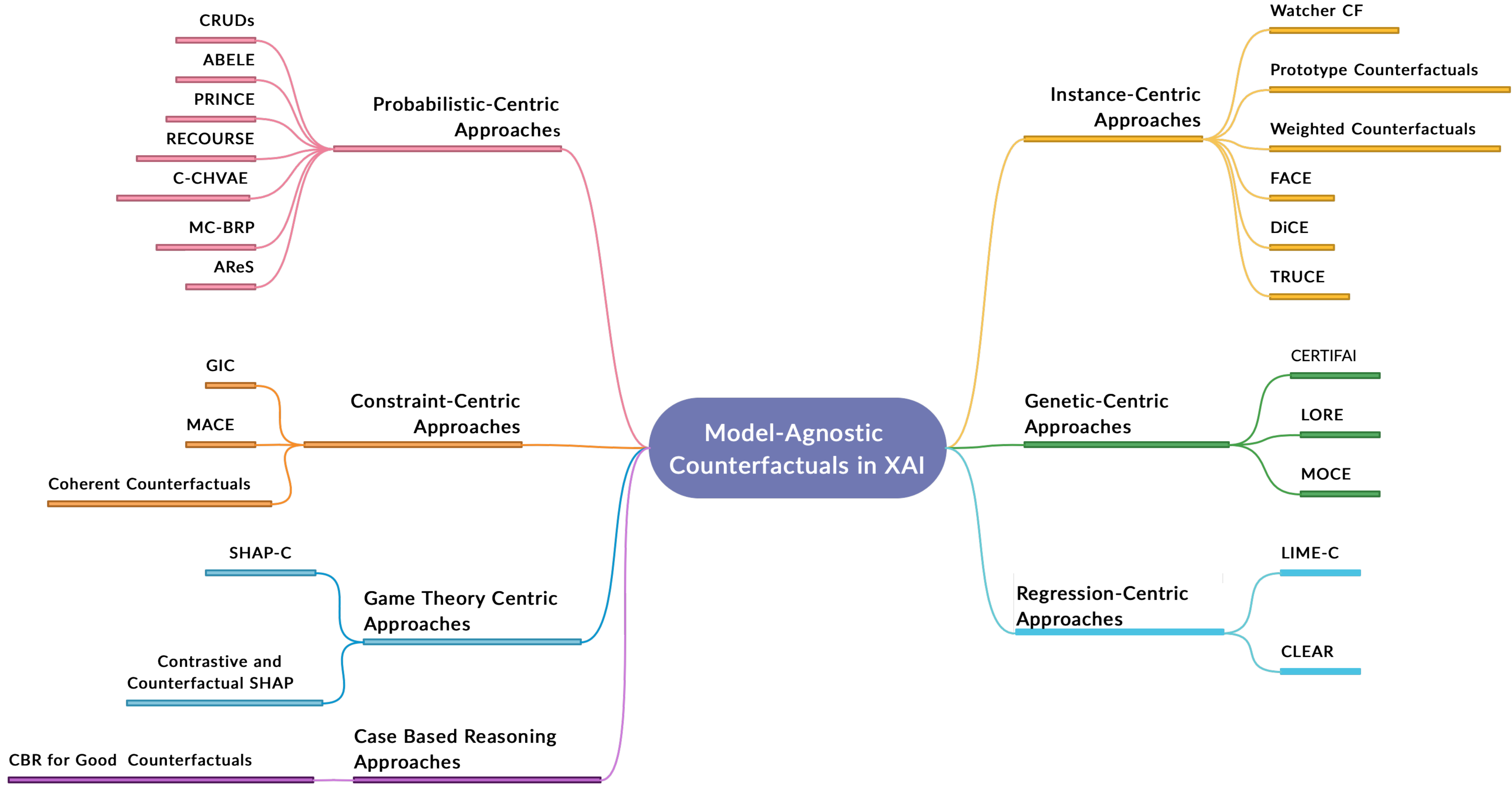}
    }
    \caption{Proposed Taxonomy for model-agnostic counterfactual approaches for XAI.}
    \label{fig:counterfactuals_taxonomy}
\end{figure}

The systematic literature review contributed to developing a new taxonomy for the model-agnostic counterfactual approaches for XAI. Throughout the review process, we noticed that many algorithms derived from similar theoretical backgrounds. In total, we analysed 26 algorithms. We created a set of six different categories representing the \textit{"master theoretical algorithm"}\citep{Domingos17} from which each algorithm derived. These categories are (1) instance-centric approaches, (2) constraint-centric approaches, (3) genetic-centric approaches, (4) regression-centric approaches, (5) game theory-centric Approaches, (6) Case-Based Reasoning Approaches, and (7) Network-Centric approaches. Figure~\ref{fig:counterfactuals_taxonomy} presents the proposed taxonomy as well as the main algorithms that belong to each category.

\begin{itemize}
	\item  \textbf{Instance-Centric.} 
	Correspond to all approaches that derive from the counterfactual formalism proposed by \citet{Lewis73counterfactual} and \citet{Wachter18}. These approaches are based on random feature permutations and consist of finding counterfactuals close to the original instance by some distance function. Instance-centric algorithms seek novel loss functions and optimization algorithms to find counterfactuals. Thus, they are more susceptible to fail the plausibility, \textcolor{blue}{the feasibility}, and the diversity properties, although some instance-centric algorithms incorporate mechanisms in their loss functions to overcome these issues.

    \textcolor{blue}{\item \textbf{Constraint-Centric.}} 
    Corresponds to all approaches that are modeled as a constraint satisfaction problem. Algorithms that fall in this category use different strategies to model the constraint satisfaction problem, such as satisfiability modulo theory solvers. The major advantage of these approaches is that they are general and can easily satisfy different counterfactuals properties such as diversity and plausibility.

    \item \textbf{Genetic-Centric.} 
    Corresponds to all approaches that use genetic algorithms as an optimization method to search for counterfactuals. Since genetic search allows feature vectors to crossover and mutate, these approaches often satisfy properties such as diversity, plausibility, and feasibility.

    \item \textbf{Regression-Centric.} 
    Corresponds to all approaches that generate explanations by using the weights of a regression model. These approaches are very similar to LIME. The intuition is that an interpretable model (in this case, linear regression) fits the newly generated data after permuting the features, and the weights of each feature presented explanations. Counterfactuals based on these approaches have difficulties satisfying several properties such as plausibility and diversity.

    \item \textbf{Game Theory Centric.} 
    Corresponds to all approaches that generate explanations by using Shapley values. These approaches are very similar to SHAP. Algorithms that fall in this approach mainly extend the SHAP algorithm to take into consideration counterfactuals. Counterfactuals based on these approaches have difficulties satisfying several properties such as plausibility and diversity.

    \item \textbf{Case-Based Reasoning.} 
    Corresponds to all approaches inspired in the case-based reasoning paradigm of artificial intelligence and cognitive science that models the reasoning process as primarily memory-based. These approaches often solve new problems by retrieving stored $cases$ describing similar prior problem-solving episodes and adapting their solutions to fit new needs. In this case, the CBR system stores good counterfactual explanations. The counterfactual search process consists of retrieving from this database the closest counterfactuals to a given query. CBR approaches can easily satisfy different counterfactual properties such as plausibility, feasibility, and diversity.
    
    \textcolor{blue}{\item \textbf{Probabilistic-Centric.} }
    \textcolor{blue}{Corresponds to approaches that model the counterfactual generation problem as a probabilistic problem. These approaches often consider random walks, Markov sampling, variational autoencoders to learn efficient data codings in an unsupervised manner, or probabilistic graphical models (PGMs). Approaches based on PGMs have the potential to satisfy the causality framework proposed by \citet{Pearl09}. }
    %Counterfactuals based on these approaches have difficulties satisfying several properties such as plausibility and diversity. So}
\end{itemize}
 
\section{Counterfactual Approaches to Explainable AI: The Algorithms}\label{sec:counterfactuals_algo}
In this systematic review, we found 18 model-agnostic XAI counterfactual algorithms. We analyzed each algorithm in-depth and classified them according to the different properties presented in Section~\ref{sec:properties}. We also classified them in terms of their applications, either for classification/regression problems and the supporting data structures. We complemented the analysis with the information of whether the algorithm is publicly available. Table~\ref{tab:algo_summary} presents a classification of collected model-agnostic counterfactual algorithms for XAI based on different properties, theoretical backgrounds, and applications.

In the following sections, each algorithm of Table \ref{tab:algo_summary} is analysed relatively to its grounding \textit{theoretical master algorithm}.

\subsection{Instance-Centric Algorithms}

In this section, we summarize the algorithms that we classified as instance-centric using the proposed taxonomy. By definition, these algorithms are very similar, diverging primarily on the loss function description with the corresponding optimization algorithm and the distance function specification.

\begin{itemize}
    \item  \textbf{WatcherCF by \citet{Wachter18}.} 
    WatcherCF correspond to one of the first algorithms in model-agnostic counterfactuals for XAI. They extend the notion of a minimum distance between datapoints that was proposed initially by \citet{Lewis73counterfactual}. The goal is to find a counterfactual $x'$ as close as possible to the original point $x_i$ as possible such that a new target $y'$ (the counterfactual) is found.

    \begin{itemize}
        \item \textbf{Loss function.}  
        The loss function takes as input the data instance to be explained, $x$, the counterfactual candidate, $x'$, and a parameter $\lambda$, that balances the distance in the prediction (first term) against the distance in feature values (second term)~\citep{Molnar18}. The higher the value of $\lambda$, the closer the counterfactual candidate, $x'$, is to the desired outcome, $y'$. Equation~\ref{eq:watcher_loss} presents the loss function and respective optimization problem proposed by~\citet{Wachter18}. \textcolor{blue}{The authors argue that the type of optimiser is relatively unimportant since most optimisers used to train classifiers work in this approach.}
        \begin{equation}
            \begin{split}
                \mathcal{L}(x, x', y', \lambda) = \lambda \left( f(x') - y'\right)^2 + d(x,x') \\
                arg~\underset{x'}{min}~\underset{\lambda}{max} \mathcal{L}(x, x', y', \lambda)~~~~~~~~~~~~~~~~~
            \end{split}
            \label{eq:watcher_loss}
        \end{equation}
        \item \textbf{Distance function.} 
        \textcolor{blue}{Although the choice of optimiser does not impact the search for counterfactuals, the choice of the distance function does.} \citet{Wachter18} argue that the $L_1$-norm normalized by the inverse of the median absolute deviation of feature $j$ over the dataset is one of the best performing distance functions because it ensures the sparsity of the counterfactual candidates.  Equation~\ref{eq:watcher_distance} presents the distance function used in their loss function.
            \begin{equation}
                \begin{split}
                     d(x, x') = \sum^{p}_{j=1} \frac{|x_j - x'_j|}{MAD_j}, \text{~ where~~~~~~~~~~~~~~~~~~~~~~~~~~~~~~~~~~~~~~~~}\\
                     MAD_j = median_{ i\in	\{1,\dots,n\}}  \left|x_{i,j} - median_{l\in \{1,\dots,n\}}(x_l,j)\right|
                \end{split}
            \label{eq:watcher_distance} 
            \end{equation}
        
        \item \textbf{Optimization algorithm:} The Adam Gradient descent algorithm is used to minimize Equation~\ref{eq:watcher_loss}.
    \end{itemize}

    \item \textbf{Prototype Counterfactuals by \citet{Looveren2019}.}  
    \textcolor{blue}{Prototype Guided Explanations consist of adding a prototype loss term in the objective result to generate more interpretable counterfactuals. The authors performed experiments with two types of prototypes: an encoder or k-d trees, which resulted in a significant speed-up in the counterfactual search ad generation process \citep{Looveren2019}}.

    \begin{itemize}
        \item \textbf{Loss function}: The loss function consists in two different steps: (1) guide the perturbations $\delta$ towards an interpretable counterfactual $x_{cf}$ which falls in the distribution of counterfactual class $i$, and (2) accelerate the counterfactual searching process. This is achieved through Equation~\ref{eq:loss_proto},
        \begin{equation}
            Loss = c.L_{pred} + \beta.L_1 + L_2 + L_{AE}+L_{proto}.
            \label{eq:loss_proto}
        \end{equation}
        The $L_{pred}$ measures the divergence between the class prediction probabilities, $L_1$ and $L_2$ correspond to the elastic net regularizer, $L_{AE}$ represents an autoencoder loss term that penalizes out-of-distribution counterfactual candidate instances (which can lead to uninterpretable counterfactuals). Finally, $L_{proto}$ is used to speed up the search process by guiding the counterfactual candidate instances towards an interpretable solution.  
        
        \item \textbf{Distance function.}
        \citet{Looveren2019} use the $L_2$-norm to find the closest encoding of perturbed instances, $ENC(x + \delta)$ of a data instance, $x$, to its prototype class, $proto_i$. This is given by Equation~\ref{eq:garth_dist}.
        
        \begin{equation}
            L_{proto} = \theta \left| \left| ENC(x + \delta) - proto_i \right| \right|_2^2
        \label{eq:garth_dist} 
        \end{equation}
        
        \item \textbf{Optimization function}: 
        \citet{Looveren2019} adopted a fast integrative threshold algorithm (FISTA) which helps the perturbation parameter $\delta$ to reach momentum for $N$ optimization steps. The $L1$ regularization has been used in the optimization function.
\end{itemize}

    %%%%%
    %%%% WEIGHTED COUNTERFACTUALS
    \item \textbf{Weighted Counterfactuals by \citet{Grath2018}.}
    Weighted counterfactuals extend WatcherCF approach in two dimensions by proposing: (1) the concepts of positive and weighted counterfactuals, and (2) two weighting strategies to generate more interpretable counterfactual, one based on global feature importance, the other based on nearest neighbours. 
    
     Traditional counterfactuals address the question \textit{why my load was not granted?} through a hypothetical \textit{what-if} scenario. On the other hand, when the desired outcome is reached, positive counterfactuals address the question \textit{how much was I accepted a loan by?} 
    
    \begin{itemize}
        \item \textbf{Loss function.} The weighted counterfactuals are computed in the same way as the WatcherCF~\citep{Wachter18} as expressed in Equation~\ref{eq:watcher_loss}.
        
        \item \textbf{Distance function.} The distance function used to compute weighted counterfactuals is the same as in WatcherCF~\citep{Wachter18}, with the addition of a weighting parameter $\theta_j$,
        \begin{equation}
            d(x,x') = \sum_{j=1} \frac{\left| x_j - x_j'\right|}{MAD_j}\theta_j.
        \end{equation}
        
        \item \textbf{Optimization algorithm.} 
        While \citet{Wachter18} used gradient descent to minimize the loss function,~\citet{Grath2018} used the Nelder-Mead algorithm, which was initially suggested in the book of~\citet{Molnar18} and is used to find the minimum of a function in a multidimensional space. The Nelder-Mead algorithm is a better algorithm to deal with the $L_1$-norm since it works well with nonlinear optimization problems for which derivatives may not be known. 
    \end{itemize}

    Experiments conducted by~\citet{Grath2018} showed that weights generated from feature importance lead to more compact counterfactuals and consequently offered more human-understandable interpretable features than the ones generated by nearest neighbours.
  
    %%%%%
    %%%% FACE
    \item \textbf{Feasible and Actionable Counterfactual Explanations \textcolor{blue}{(FACE)}} by \citet{Poyiadzi20}.
    
    FACE aims to build coherent and feasible counterfactuals by using the shortest path distances defined via density-weighted metrics. This approach allows the user to impose additional feasibility and classifier confidence constraints naturally and intuitively. Moreover, FACE uses Dijkstra's algorithm to find the shortest path between existing training datapoints and the data instance to be explained~\citep{Verma20}.
    
    \textcolor{blue}{Under this approach, feasibility refers to the search of a counterfactual that does not lead to \textit{paradoxical interpretations}. For instance, low-skilled unsuccessful mortgage applicants may be told to double their salary, which may be hard without first increasing their skill level. This may render to counterfactual explanations that are impractical and sometimes outright offensive \citep{Poyiadzi20}}.
    
    \begin{itemize}
        \item \textbf{Main Function.} 
        The primary function of FACE's algorithm is given by Equation~\ref{eq:face_main}, where $f$ corresponds to a positive scalar function and $\gamma$ is a function that connects the path between a data instance $x_i$ and a counterfactual candidate instance $x_j$. 
            \begin{equation}
                \begin{split}
                \hat{\mathcal{D}}_{f,\gamma} =\sum_i f_p\left( \frac{\gamma(t_{i-1}) + \gamma(t_i)}{2} \right) . \left| \left|.\gamma(t_{i-1}) - \gamma( t_i )  \right| \right|, \text{where } \\
                    \hat{\mathcal{D}}_{f,\gamma} = \int_\gamma f(\gamma(t)) . \left| \gamma'(t) \right|dt.~~~~~~~~~~~~~~~~~~~~~~~~~~~~~~~~~~~~~~~~~~~~~
                \end{split}
            \label{eq:face_main}
            \end{equation}
            
            When the partition $\mathcal{\hat{D}}_{f,\gamma}$ converges, ~\citet{Poyiadzi20} suggest, for a given threshold $\epsilon$, using weights of the form
            \begin{equation}
                \begin{split}
                    w_i,j = f_p \left( \frac{x_i + x_j}{2} \right) . \left| \left| x_i - x_j \right| \right|, \\
                    \text{when~~~~~}   \left| \left| x_i - x_j \right| \right| \leq \epsilon.
                \end{split}
                \label{eq:face_weights}
            \end{equation}
            
            The $f$-distance function is used to quantify the trade-off between the path length and the density in the path. This can subsequently be minimized using Dijkstra's shortest path algorithm by approximating the $f$-distance using a finite graph over the data set.
        
        \item \textbf{Distance Function.}  
        \citet{Poyiadzi20} used the $L_2$-norm in addition to Dijkstra's algorithm to generate the shortest path between a data instance $x_i$ and a counterfactual candidate instance $x_j$. 
        
        \item \textbf{Optimization Function.} 
        \citet{Poyiadzi20} suggested three approaches that can be used to estimate the weights in Equation~\ref{eq:face_weights}:
            \begin{equation}
                \begin{split}
                    w_{i.j} = f_{\hat{p}}\left( \frac{x_i + x_j}{2} \right) . \left| \left| x_i - x_j  \right| \right| ~~~~~~~~~~~~~~~~~~~~~~~~~\\
                    w_{i.j} = \tilde{f}\left( \frac{r}{\left| \left| x_i + x_j \right| \right|}\right) . \left| \left| x_i - x_j  \right| \right|, ~~~ r = \frac{k}{N . \eta_d}\\
                    w_{i.j} = \tilde{f}\left( \frac{\varepsilon^d}{\left| \left| x_i + x_j \right| \right|}\right) . \left| \left| x_i - x_j  \right| \right|
                    ~~~~~~~~~~~~~~~~~~~\\
                \end{split}
            \end{equation}
        The first equation requires using Kernel Density Estimators to allow convergence, the second requires a k-NN graph construct, and the third equation requires $\epsilon$-graphs.
        In their experiments,~\citet{Poyiadzi20} found that the third weight equation together with $\epsilon$-graphs generated the most feasible counterfactuals.
    \end{itemize}
    
    %%%%%
    %%%% DICE
    \item \textbf{Diverse Counterfactual Explanations (DiCE)  by \citet{Mothilal20}.}
    DICE is an extension and improvement of the WatcherCF~\citep{Wachter18} throughout different properties: Diversity, proximity, and sparsity. DiCE generates a set of diverse counterfactual explanations for the same data instance $x$, allowing the user to choose counterfactuals that are more understandable and interpretable. Diversity is formalized as a determinant point process, which is based on the determinant of the matrix containing information about the distances between a counterfactual candidate instance and the data instance to be explained.

        \begin{itemize}
            \item \textbf{Loss Function.} 
            In DiCE, the loss function is presented in Equation~\ref{eq:dice}, and is given by a linear combination of three components: (1) a hinge loss function that is a metric that minimizes the distance between the user prediction $f(.)'$ for $c_is$ and an ideal outcome $y$, $loss(f(c_i),y)$, (2) a proximity factor, which is given by a distance function, and (3) a diversity factor $dpp\_diversity(c_1, ..., c_k)$. 
            \begin{equation}
              \begin{split}
                C(x) = \underset{c_1,...,c_k}{\operatorname{arg min}} \frac{1}{k}\sum_{i=1}^{k} yloss(f(c_i),y) + \frac{\lambda_1}{k}\sum_{i=1}^{k} dist(c_i,x) \\
                - \lambda_2 \ ddp\_ diversity(c_1,...,c_k)~~~~~~~~~~~~~~~~~
                \end{split}
                \label{eq:dice}
           \end{equation}
            
            \item \textbf{Distance Function.} 
            DiCE uses the $L_1$-norm normalized by the inverse of the median absolute deviation of feature $j$ over the dataset just like in $b$-counterfactual~\citep{Wachter18}.
            
            \item \textbf{Optimization Function.} Gradient descent is used to minimize Equation~\ref{eq:dice}.
        \end{itemize}

    %%%%
    %%% TRUCE
    \item \textbf{Unjustified Counterfactual Explanations (TRUCE) by \cite{Laugel18,Laugel19,Laugel2019unjustified}}
        TRUCE consists in approaching the problem of determining the minimal changes to alter a prediction by proposing an inverse classification approach~\citep{Laugel18}. The authors present the \textit{Growing Spheres} algorithm, which consists of identifying a close neighbour classified differently through the specification of sparsity constraints that define the notion of \textit{closeness}.
        
        \begin{itemize}
        \item \textbf{Loss Function.} 
        To simplify the process for reaching the closest desirable feature, \cite{Laugel19} presented a formalization for binary classification by finding an observation value $e$, then classified it into a different class other than $x$. For instance, $f(e) \neq f(x)$, indicates that the observation has been classified into the same class as $x$, and a desirable feature has been found if it is classified to the other class. For the next step, a function has been defined $c : X \times X \rightarrow R^+$ such that $c(x, e)$ is the cost of moving from observation $x$ to enemy $e$. 

             \begin{equation}
                \begin{split}
                  e^* =  arg \underset{e\in X}{min} \{c(x,e)~|~f(e) \neq f(x)\}~~~\text{with}  \\ 
                  c(x,e) = \left| \left| x-e \right|\right|_2 + \gamma\left| \left|x-e\right|\right|_0
                  \label{eq:truce}
                \end{split}
            \end{equation}    
        %%%%%
        
        \item \textbf{Distance Function.}
        \textcolor{blue}{Equation \ref{eq:truce} consists in the minimisation of a cost function under the constraint that the observation $e$ is classified into the same class as $x$. This cost function is defined as a weighted linear equation consisting of $L_2$-norm and the $L_0$-norm between the observation $e$ and the class $x$. The $L_2$-norm computes the proximity between $x$ and $e$, while the $L_0$-norm is used as a weighted average to guarantee that the explanation is human-interpretable.}
        
        \item \textbf{Optimization Function.}
        The authors used the proposed growing sphere algorithm to handle as the optimiser. The algorithm applies a greedy method to find the closest feature in all possible directions until the decision boundary is reached. This means that the $L_2$-norm was successfully minimized. The minimum feature change is also addressed through the minimisation of the $L_0$-norm.
        
        \textcolor{blue}{In a later work, the authors proposed to distinguish between justified and unjustified counterfactual explanations \citep{Laugel19,Laugel2019unjustified}. In this sense, unjustified explanations refer to counterfactuals that result from artifacts learned by an interpretable post-hoc model and do not represent the ground truth data. On the other hand, justified explanations refer to counterfactual explanations according to the ground truth data.}
        
    \end{itemize}
    
\end{itemize}

\subsection{Constraint-Centric Approaches}

In this section, we summarize the algorithms that we classified as constraint centric using the proposed taxonomy. 

\begin{itemize}
    %%%%
    %%% MACE
    \item  \textbf{Model-Agnostic Counterfactual Explanations for Consequential Decisions (MACE) by \citet{Karimi2020mace}}
    MACE maps the problem of counterfactual explanation search into a satisfiability modulo theory (SMT) model. 
    MACE receives as input a sequence of satisfiability problems expressing the predictive model, $f$, the distance function, $d$, and the constraint functions. The algorithm's goal is to map these sequences into logical formulae and verify if there is a counterfactual explanation that satisfies a distance smaller than some given threshold. The constraints that are taken into consideration in this approach are \textit{plausibility} and \textit{diversity}. This is achieved in the following way. Given the counterfactual logical formula $\phi_{CF_f(\hat{x})}$, the distance formula $\phi_{d,\hat{x}}$, constraints formula $\phi_{g,\hat{x}}$, and a threshold $\epsilon$, they are combined into the counterfactual formula, $\phi_{\hat{x}, \delta}(x)$, given by
    \begin{equation}
        \phi_{\hat{x}, \delta}(x) = \phi_{CF_f(\hat{x})} (x) \land \phi_{d,\hat{x}} \land \phi_{g,\hat{x}},
        \label{eq:mace}
    \end{equation}
    and used as input for a SMT solver, $SAT( \phi_{\hat{x}, \delta}(x) )$, which will find counterfactuals that will satisfy the conditions with a distance smaller than $\epsilon$. MACE is a general algorithm that supports any $L_p$-norm as a distance function, as well as any number of constraints. 
    
    \textcolor{blue}{MACE was able to not only achieve high plausibility (what the authors define as \textit{coverage}) but was also able to generate counterfactuals at more favorable distances than existing optimization-based approaches \citep{Karimi2020mace}.}
 \end{itemize} 

    %%%% TO REVISE
\begin{itemize}

    %%%%
    %% COHERENT COUNTERFACTUALS
    \item \textbf{Coherent Counterfactuals by \citet{Russell19}}
    
   This approach focuses on generating diverse counterfactuals based on "mixed polytope" methods to handle complex data with a contiguous range or an additional set of discrete states. \citet{Russell19} created a novel set of criteria for generating diverse counterfactuals to integrate them with the mixed polytope method and to map them back into the original space. Before achieving the two targets (coherence and diversity), \citet{Russell19} firstly offers a solution on generating a counterfactual which is similar to \citet{Wachter18}'s counterfactuals by finding the minimum change in the counterfactual candidate that would lead to a change in the prediction (using the $L_1-norm$). Then, the author proposes the mixed polytope as a novel set of constraints. The program uses an integer programming solver and receives a set of constraints in order to find coherent and diverse counterfactuals efficiently. 

    %%%%%%
    %%% GIC - GENERALISED INVERSE CLASSIFICATION
    \item \textbf{Generalized Inverse Classification by \citet{Lash2017}}
     
    Inverse classification is similar to the sub-discipline of sensitivity analysis, which examines the impact of predictive algorithm input on the output. \cite{Lash2017} proposed Inverse Classification framework, which mainly focuses on optimizing the generation of counterfactuals through a process of perturbing an instance. This task is achieved by operating on features and tracking each change that leads to an individual cost. Every change in perturbing an instance is subject to happen within a certain level of cumulative change. 
    
     To assess the capability of GIC, \cite{Lash2017} applied five methods, including three heuristic-based methods to solve the generalized inverse classification problem, a hill-climbing + local search (HC+LS), a genetic algorithm (GA), and a genetic algorithm + local search (GA+LS). The other two methods are sensitivity analysis-based methods such as Local Variable Perturbation–Best Improvement (LVP-BI) and Local Variable Perturbation–First Improvement (LVP-FI). These five algorithms were tested to examine the average likelihood of test instances conforming to a non-ideal class over varying budget constraints. The final result showed that LVP-FI outperforms all other methods, while LVP-BI is comparable to GA and GA+LS. HC+LS has the worst performs. 
    
\end{itemize}

\subsection{Genetic-Centric Approaches}

In this section, we summarize the algorithms that we classified as genetic-centric using the proposed taxonomy. 

\begin{itemize}
    %%%%%
    %%%% LORE
    \item  \textbf{Local Rule-Based Explanations of Black Box Decision Systems by \citet{Guidotti18}}
    
    This approach attempts to provide interpretable and faithful explanations by learning a local interpretable predictor on a synthetic neighborhood generated by a genetic algorithm. Explanations are generated by decision rules that derive from the underlying logic of the local interpretable predictor.
    
    LORE works as follows. Given a black-box predictor and a local counterfactual instance, $x$, with outcome $y$, an interpretable predictor is created by generating a balanced set of neighbor instances of the given instance $x$ using an ad-hoc genetic algorithm. The interpretable model used to fit the data corresponds to a decision-tree from which sets of counterfactual rules can be extracted as explanations.

    The distance function used in this algorithm is given by Equation~\ref{eq:lore}. 
    The fitness function used corresponds to the distance of $x$ to a generated counterfactual candidate $z$,  $d(x, z)$.
    This algorithm also considers the mixed types of features by a weighted sum of the simple matching coefficient for categorical features, and by using the $L_2$-norm to normalize the continuous features. Assuming $h$ corresponds to categorical features and $m-h$ to continuous ones, then the distance function is given by
        \begin{equation}
            d(x,z) = 
\frac{h}{m} . SimpleMatch(x, z) + \frac{m-h}{m} . NormEuclid(x, z).
            \label{eq:lore}
        \end{equation}
    
    %%%%%
    %%%% CERTIFAI
    \item \textbf{Counterfactual Explanations for Robustness, Transparency, Interpretability, and Fairness of Artificial Intelligence models (CERTIFAI) by~\citep{Sharma19}}
    
    CERTIFAI is a custom genetic algorithm based explanation with several strengths, including the capability of evaluating the robustness of a machine learning model (CERScore) and assessing fairness with linear and non-linear models and any input form (from mixed tabular data to image data) without any approximations to or assumptions for the model. 
    
    Establishing a CERTIFAI framework comes with several steps that start by creating an original genetic framework, selecting a distance function, and improving counterfactuals with constraints. A custom genetic algorithm was made in the first stage by considering $f$ as a classifier for a black-box model, and an instance $x$ as an input. In this formalist, consider $c$ as the counterfactual candidate instance of $x$ and $d(x,c)$ the distance between them. The distance function used is the $L_1$-norm normalized by the median absolute deviation (MAD), as proposed by \citet{Wachter18}. The goal is to minimize the distance between $x$ and $c$ by applying a genetic algorithm.
    
    Given a variable $W$, we define the space from which individuals can be generated, to ensure feasible solutions. By taking $n$ dimensions as input, 
    multiple constraints need to be created to match with continuous, categorical, and immutable features. For instance, $W_1, W_2...,W_n$ is defined for continuous feature constraints as $W_i \in {W_{imin, W_{imax}}} $, and $W_i \in {W_1,W_2, ..., W_j}$ is for categorical variables. Finally, a feature $i$ for an input $x$ should be mutated by setting $W_i = X_i$.

    The robustness and fairness of the population of the generated counterfactuals are given by
    \begin{equation}
        CERScore(model) = \frac{E}{x}[d(x, c^*)].
    \end{equation}
    Fairness ensures that the solutions generated contain different counterfactuals with multiple values of an unchangeable feature (e.g., gender, race).

    %%%%%
    %%% MOCE
    \item \textbf{Multi-Objective Counterfactual Explanations (MOCE) by \citep{Dandl2020}.}
    This approach consists of a multi-objective counterfactual explanation algorithm which translates the counterfactual search into a multi-objective optimization problem based on genetic algorithms. This approach brings the benefit of providing a diverse set of counterfactuals with a variety of trade-offs between the proposed objectives and maintains diversity in feature space at the same time. 
  
    \citet{Dandl2020} proposed a four-objective loss equation to generate an explanation:
    \begin{equation}
        L(x, x', y', X^{obs}) = (o_1(f^\wedge(x',y'), o_2(x, x'), o_3(x, x'), o_4(x', X^{obs}))).
    \end{equation}
    In the proposed equation, the four objectives $o_1 to o_4$ represent one of the four criteria:  Objective 1, $o_1$, focuses on generating the closest possible result from a prediction of counterfactual $x'$ to the desired prediction $y'$. It minimizes the distance between $f(x')$ and $y'$, and calculates it through the $L_11$-norm. Objective 2 states that the ideal counterfactual should be as similar as possible to instance $x$. It quantifies the distance between $x'$ and $x$ using Grower's distance. Objective 3, $o_3$ is used to calculate the sparse feature changes through $L_0$-norm. This norm is necessary because Grower distance can handle numerical and categorical features but cannot count how many features were changed. Finally, Objective 4 states that the ideal counterfactual should have similar feature value combinations as the original data point. The solution is to measure how "likely" a data point uses the training data, $Xobs$.
    
  \citet{Dandl2020} used the Nondominated Sorting Genetic Algorithm or short NSGA-II, which is a nature-inspired algorithm and applies Darwin's law of the survival of the fittest and denote the fitness of a counterfactual by its vector of objectives values ($o1,o2,o3,o4$), this solution helps to produce a fitter counterfactual result by showing the lower counterfactuals four objectives.

\end{itemize}

\subsection{Regression-Centric Approaches}

In this section, we summarize the algorithms that we classified as regression-centric using the proposed taxonomy:

\begin{itemize}
    \item \textbf{Counterfactual Local Explanations via Regression (CLEAR) by \citet{White19}}:
    This method aims to provide a counterfactual explained by regression coefficients, including interaction terms, and significantly improve the fidelity of the regression. \cite{White19} firstly generates boundary counterfactual explanations, which state minimum changes necessary to flip a prediction’s classification, then builds local regression models using the boundary counterfactual to measure and improve the fidelity of its regressions. . 
    
    CLEAR proposes a \textit{counterfactual fidelity error}, $CFE$ , which is based on the concept of $b$-perturbation. It compares each $b$-perturbation with an estimation of that value, $estimated b$-perturbation, which is calculated by a local regression. $CFE$ is given by
    \begin{equation}
        CFE = \left|estimated~b-perturbation - b-perturbation \right|.
    \end{equation}
    The generation of counterfactuals in CLEAR has the following steps: (1) Determine $x$’s boundary perturbations; (2) Generate labeled synthetic observations; (3) Create a balanced neighborhood dataset; (4) Perform a stepwise regression on the neighborhood dataset, under the constraint that the regression curve should go through $x$; (5)  Estimate the $b$-perturbations; (6) Measure the fidelity of the regression coefficients; (7) Iterate until the best explanation is found. 

   \cite{White19} compared the performance in terms of fidelity of CLEAR and LIME. The result showed that regressions have significantly higher fidelity than LIME in five case studies.

    %%%%%%
    %%%% LIME-C
    \item \textbf{Local Interpretable Model-Agnostic Explanations-Counterfactual (LIME)-C by \citet{Ramon20}}. 
    LIME-C is a hybrid solution which connects additive feature attribution explanations (like in LIME) with counterfactuals. The motivation for this hybrid solution starts from an assumption which states that if the importance-rankings of the features are sufficiently accurate, it may be possible to compute counterfactuals from them. 
    
    Additive feature attribution methods use an explanation model $g$ as an interpretable approximation of the trained classification model $C$, which can be represented by a linear model \citep{Ramon20}:
    \begin{equation}
        g(x') = \phi_0 + \displaystyle\sum_{j=1}^{m} \phi_j x'_j
        \label{eq:linec}
    \end{equation}
    In Equation \ref{eq:linec}, $x'_j \in {0, 1}$ is the binary representation of $x'_j$ (where $x'_j$ is 1, if $x'_j$ is non-zero, else it equals 0), $m$ is the number of features of instance $x$, and $\phi_0, \phi_j \in \mathcal{R}$. To generate a ranked list of counterfactual, the authors used the linear algorithm for finding counterfactuals (SEDC) proposed by \cite{Martens14}.
    
    \citet{Ramon20} points out that this method is stable and effective for all data and models, and even for very large data instances that require many features to be removed for a predicted class change, LIME-C computes counterfactuals relatively fast.

    %%%%
    %%%% SEDC
    \item \textbf{Search for Explanations for Document Classification (SEDC) by \citet{Martens14}}

    This approach was proposed in the domain of information retrieval back in 2014. It consists of generating explanations for the user's understanding of the system and also for model inspection. 
    SEDC was one of the first works that used \citet{Lewis73counterfactual} definition of counterfactual in an algorithm that provides minimal sets of words as explanations, such as removing all words within this set from the document that changes the predicted class from the class of interest.

  SEDC outputs minimum-size explanations for linear models by ranking all words appearing in the document through the product $\beta_{jX_{ij}}$, where $\beta_j$ is the linear model coefficient. The explanation with the top-ranked words is an explanation of the smallest size, and therefore a counterfactual. Cosine-similarity is used to measure the proximity between a document and the counterfactual document candidate.
    
\end{itemize}

\subsection{Game Theory Centric Approaches}

In this section, we summarize the algorithms that we classified as Game Theory Centric using the proposed taxonomy.

\begin{itemize}
    \item \textbf{SHAP Counterfactual (SHAP-C) by \cite{Ramon20}}
    
    SHAP-C is a hybrid algorithm that combines Kernel SHAP~\citep{Lundberg17} with SEDC~\citep{Martens14}.
    
    SHAP-C works as follows. Given a data point and a black-box predictive model, first, we compute the respective Shapley values using kernel SHAP. The algorithm ranks the most important features by their respective SHAP values and adds these features to a set called the \textit{Evidence Counterfactual}. The algorithm then proceeds similarly as in SEDC: the most important features are perturbed so that a minimum set of perturbations is found to flip the prediction of the datapoint. This \textit{evidence counterfactual} is returned as the explanation.
     
     %%%%%
     %%%% 
    \item \textbf{Contrastive and Counterfactual SHAP (SHAP-CC) by \cite{Rathi2019}.}
    
   SHAP-CC attempts to generate partial post-hoc contrastive explanations with a corresponding counterfactual. \citet{Rathi2019} used a P-contrastive methodology for generating contrastive explanations that would allow the user to understand why a particular feature is important, and why another specific feature is not. 
    
    The main idea of this explanation is considered a P-contrast question which is  equivalent to the format \textit{"Why [predicted-class]"} instead of \textit{[desired class]?}. To answer these questions, \citet{Rathi2019} explanation computed the Shapely values for each of the possible target classes where a negative Shapely value indicates the features that have negatively contributed to the specific class classification. \citet{Rathi2019} generate a "Why P not Q" explanation by breaking it down into two questions: "why P?" and "Why not Q." The positive and negative Shapley values are given as an answer to these questions, respectively. Furthermore, a contrastive and counterfactual explanation is given in terms of natural language to the user.

    \end{itemize}

\subsection{Case-Based Reasoning Approaches}

In this section, we summarize the algorithms that we classified as Case-Based Reasoning using the proposed taxonomy.

\begin{itemize}
    \item\textbf{Case-based reasoning (CBR) for Good counterfactual by \cite{Keane2020}}
    
    CBR for good counterfactuals uses a case-based system where examples of good counterfactuals are stored. By good counterfactuals, \citet{Keane2020} understands as counterfactuals that are sparse, plausible, diverse, and feasible. The authors also introduce the \textit{explanatory coverage} and \textit{counterfactual potential} as properties of CBR systems that can promote good counterfactuals.
    
    \textcolor{blue}{In their algorithm, \citet{Keane2020} refer to the pairing of a case and its corresponding good counterfactual as an \textit{explanation case} or $XC$. The goal is to generate good counterfactuals by retrieving, reusing, and revising a nearby explanation case by taking the following steps: (1) identify the $XC$ that is most similar to the query datapoint $p$. In other words, $XC$ corresponds to the explanatory coverage between two data points, $xc(x,x')$; (2) for each of the features matched in $xc(x,x')$; we map these features from $p$ to the new generated counterfactual $p'$. In the same way, we add to $p'$ the most different features in $xc(x,x')$. This procedure guarantees the diversity of the counterfactuals; (3) through the definition of a counterfactual, $p'$ needs to provide a prediction contrary to the one by $p$. This implies that $p'$ needs to go through the predictive black-box and returned to the user if the prediction flips; (4) otherwise, an adaptation step to revise the
values of the different features in $p'$ are applied until there is a change of class.}

\end{itemize}

\textcolor{blue}{\subsection{Probabilistic-Centric Approaches}}

\textcolor{blue}{In this section, we summarize the algorithms that we classified as Probabilistic-Centric using the proposed taxonomy:}

\begin{itemize}
    %%%
    %%% PRINCE: PROVIDER SIDE INTERPRETABILITY WITH COUNTERFACTUAL EXPLANATIONS IN RECOMMENDER SYSTEMS
    \item\textbf{Provider-side Interpretability with Counterfactual Explanations in Recommender Systems (PRINCE) proposed by \citet{Ghazimatin20}}
    
    \textcolor{blue}{This approach aims to detect the actual cause of a recommendation by a heterogeneous information network (HIN) with users, items, reviews, and categories. It identifies a small set of a user's actions by removing actions that would lead to replacing the recommended item with a different item.}
    
    \textcolor{blue}{This approach provides user explanations by display what they can do to receive more relevant recommendations to the users. Personalized PageRank(PPR) scores were chosen as the recommender model to create a heterogeneous knowledge network. PRINCE is based on a polynomial-time algorithm for searching the search space for subsets of user behaviors that could lead to a recommendation. By adopting the reverse local push algorithm to a dynamic graph environment, the algorithm efficiently computes PPR contributions for groups of actions about an object.}
    
    \textcolor{blue}{Experiments performed by \citet{Ghazimatin20} using data from Amazon and Goodreads showed that simpler heuristics fail to find the best explanations. PRINCE, on the other hand, can guarantee optimality, since it outperformed baselines in terms of interpretability in user studies.}

    %%%
    %%% C-CHVAE COUNTERFACTUAL CONDITIONAL HETEROGENEOUS AUTO-ENCODER 
    \item \textcolor{blue}{\textbf{Counterfactual Conditional Heterogeneous Autoencoder (C-CHVAE) by \citet{Pawelczyk2020CCHVAE,Pawelczyk2019}}}
    
    \textcolor{blue}{C-CHVAE uses Variational Autoencoders (VAE) to search for faithful counterfactual, which consists in counterfactuals that are not local outliers, and that are \textit{connected} to regions with significant data density (similar to the notion of feasibility introduced by \citet{Poyiadzi20}). 
    Given the original dataset, the data is converted into an \textit{encoding vector} where the data is represented in a lower dimension, and each dimension represents some learned probability distribution. It is this encoder that specifies which low dimensional neighbourhood one should look for potential counterfactuals. The next steps consist in perturbing the low dimensional data and pass it through a decoder, which will reconstruct the lower dimensional potential counterfactuals into their original space. Finally, the new generated counterfactuals are given to a pre-trained black-box in order to assess whether the prediction has been altered.}
    
      \item \textcolor{blue}{\textbf{Monte Carlo Bounds for Reasonable Predictions (MC-BRP) by \citet{Lucic2020}}.
        MC-BRP is an algorithm that focuses on predictions with significant errors where Monte Carlo sampling methods generate a set of permutations of an original data point instance that result in reasonable predictions. 
   Given a local instance, $x_i$, a set of important features $\Phi^{(x_i)}$, a black-box model $f$, and an error threshold $\epsilon$, MC-BRP uses Tuckey's fence to determine outliers (predictions with high errors) for each feature of the set of important features, 
   \[ \epsilon > Q3(E) + 1.5(Q3(E) - Q1(E)), \]
   where $Q_1(E)$, $Q_3(E)$ are the first and third quartiles of the set of errors, $E$, of each feature, respectively.
    Tukey's fence will return a set of boundaries for which reasonable predictions would be expected for each of the most important features. Using these boundaries, MC-BRP generates a set of permutations using Monte Carlo sampling methods, which will be passed into the black-box $f$ to obtain a new prediction. Finally, a trend is computed based on Pearson correlation over the reasonable new predictions. The reasonable bounds for each feature are recomputed and presented to the user in a table.
   }
   
   \textcolor{blue}{
    \item \textbf{Adversarial Black box Explainer generating Latent Exemplars (ABELE) by} \citet{Guidotti20}}
    
    \textcolor{blue}{ABELE is a local model-agnostic explainer for image data that uses Aversarial AutoEncoderes (AAEs) that aim at generating new counterfactuals that are highly similar to the training data.}
    
    \textcolor{blue}{ABELE generates counterfactuals in four steps: (1) by generating a neighborhood in the latent feature space using the AAEs, (2) by learning a decision tree on the generated latent neighborhood by providing local decision and counterfactuals rules, (3) by selecting and decoding exemplars and counter-examples satisfying these rules, and (4) by extracting the saliency maps out of them.}

    \textcolor{blue}{\citet{Guidotti20} found that ABELE outperforms current state-of-the-art algorithms, such as LIME, in terms of coherence, stability, and fidelity.} 
   
    %%%%%
    %% CRUDS: COUNTERFACTUAL RECOURSE USING DISENTAGLED WORKSPACES
    \textcolor{blue}{\item \textbf{CRUDS: Counterfactual Recourse Using Disentangled Subspaces by \citet{Downs2020crud}}}
    
    \textcolor{blue}{CRUDs is a probabilistic model that uses conditional subspace variational auto encoders (CSVAEs) that extracts latent features that are relevant for a prediction. CSVAE partitions the latent space into two parts: one to learn representations that are predictive of the labels, and one to learn the remaining latent representations that are required to generate data.}
    
    \textcolor{blue}{In CRUDs, counterfactuals that target desirable outcomes are generated using CSVAEs in five major steps: (1) disentangling latent features relevant for classification from those that are not, (2) generating counterfactuals by changing only relevant latent features, filtering cuonterfactuals given constraints, and (4) summarise counterfactuals for interpretability.}

    \textcolor{blue}{\citet{Downs2020crud} evaluated CRUDS on seven synthetically generated and three real datasets. The result indicates that CRUDS counterfactuals preserve the true dependencies between the covariates in the data in all datasets except one.}
    
    %%%%%
    %% Recourse: Algorithmic Recourse Under Imperfect Causal Knowledge
    \textcolor{blue}{\item \textbf{Recourse: Algorithmic Recourse Under Imperfect Causal Knowledge  by \citet{Karimi2020nips}}}
    
    \textcolor{blue}{Traditional works on counterfactuals for XAI focus on finding the nearest counterfactual that promotes a change in the prediction to a favourable outcome. On the other hand, algorithm recourse approaches focus on generating a set of actions that an individual can perform to obtain a favourable outcome. This implies a shift from minimising a distance function to optimising a personalised cost function~\citep{Karimi2020nips}.}
    
    \textcolor{blue}{To the best of our knowledge, recourse is the only model-agnostic algorithm in the literature that attempts to use a causal probabilistic framework as proposed in \citet{Pearl09} grounded on structural causal models to generate counterfactuals, and where \textit{recourse actions} are defined as interventions of the form of $do-calculus$ operations.} 
    
    \textcolor{blue}{\citet{Karimi2020nips} highlight that it is very challenging to extract a structured causal model (SCM) from the observed data \citep{Barocas2020}, and in their work, they assume an imperfect and partial SCM to generate the recourse, which can lead to incorrect counterfactuals. To overcome this challenge, \citet{Karimi2020nips} proposed two probabilistic approaches that relax the strong assumptions of an SCM: the first one consists of using additive Gaussian noise and Bayesian model averaging to estimate the counterfactual distribution; the second removes any assumptions on the structural causal equations by computing the average effect of recourse actions on individuals. Experiment results on synthetic data showed that \textit{Recourse} was able to make more reliable recommendations under a partial SCM than other non-probabilistic approaches.}
\end{itemize}

\subsection{Summary}

We conducted a thorough systematic literature review guided by the argument to measure the causability of an XAI system. Then the system needs to be underpinned by a probabilistic causal framework such as the one proposed in~\citet{Pearl09}. We believe that counterfactual reasoning could provide human causal understandings of explanations, \textcolor{blue}{although some authors challenge this notion. \citet{Zheng2020} conducted studies to investigate whether presenting causal explanations to a user would lead to better decision-making. Their results were mixed. They found that if the user has prior and existing domain knowledge, then presenting causal information did not improve the decision-making quality of the user. On the other hand, if the user did not have any prior knowledge or beliefs about a specific task, then causal information enabled the user to make better decisions.}

We also found that the majority of the counterfactual generation approaches are not grounded on a formal and structured theory of causality as proposed by \citet{Pearl09}. Current counterfactual explanation generation approaches for XAI are based on spurious correlations rather than cause-effect relationships. This inability to disentangle correlation from causation can deliver sub-optimal, erroneous, or even biased explanations to decision-makers, as \citet{Richens2020} highlighted in his work about medical decision making. 

Table~\ref{tab:algo_summary} summarizes all the algorithms that we analysed in this section in therms of underlying theories, algorithms, applications and properties.

\begin{table}[!h]
\resizebox{\columnwidth}{!} {
\begin{tabular}{clccc|c|c|c|c|c|c|c|}
\cline{6-12}
\multicolumn{5}{c}{} &
  \multicolumn{7}{c|}{\cellcolor[HTML]{C0C0C0}\textbf{Properties}} \\ \hline
\rowcolor[HTML]{C0C0C0} 
\multicolumn{1}{|c|}{\cellcolor[HTML]{EFEFEF}\textbf{Theory / Approach}} %&Theory}} 
&
  \multicolumn{1}{c|}{\cellcolor[HTML]{EFEFEF}\textbf{Algorithms}} &
  \multicolumn{1}{c|}{\cellcolor[HTML]{EFEFEF}\textbf{Ref.}} &
  \multicolumn{1}{c|}{\cellcolor[HTML]{EFEFEF}\textbf{Applications}} &
    \multicolumn{1}{c|}{\cellcolor[HTML]{EFEFEF}\textbf{Code?}} &
  \textcolor{blue}{\textbf{Proximity}} &
  \textbf{Plausibility} &
  \textbf{Sparsity} &
  \textbf{Diversity} &
  \textbf{Feasibility} &
  \textbf{Optimization} &
  \textbf{Causal?} \\ \hline
  
  % INSTANCE CENTRIC
  \multicolumn{1}{|c|}{\cellcolor[HTML]{FFFC9E}} &
    %%%%
  %%% WATCHER COUNTERFACTUALS
  \multicolumn{1}{l|}{WatcherCF} &
  \multicolumn{1}{c|}{\citep{Wachter18}} &
  \multicolumn{1}{c|}{\begin{tabular}[c]{@{}c@{}}C\\ {[}Tab / Img{]}\end{tabular}} &
  \begin{tabular}[c]{@{}c@{}}Yes \citep{Alibi}\\ {[}Algo: CF{]}\end{tabular} &
  \begin{tabular}[c]{@{}c@{}}\greencheck\\ {[}L$_1$-norm{]}\end{tabular} &
  \crosscheck &
  \greencheck &
  \crosscheck &
  \crosscheck &
  Gradient Descent &
  \crosscheck \\ \cline{2-12}

  %%%%
  %%% PROTOTYPE COUNTERFACTUALS
\multicolumn{1}{|c|}{\cellcolor[HTML]{FFFC9E}} &
  \multicolumn{1}{l|}{\begin{tabular}[c]{@{}l@{}}Prototype \\ Counterfactuals\end{tabular}} &
  \multicolumn{1}{c|}{\citep{Looveren19}} &
  \multicolumn{1}{c|}{\begin{tabular}[c]{@{}c@{}}C\\ {[}Tab / Img{]}\end{tabular}} &
  \begin{tabular}[c]{@{}c@{}}Yes \citep{Alibi}\\ {[}Algo: CFProto{]}\end{tabular} &
  \begin{tabular}[c]{@{}c@{}}\greencheck\\ {[}L$_1$/L$_2$-norm{]}\end{tabular} &
  \greencheck &
  \begin{tabular}[c]{@{}c@{}}\greencheck\\ {[}kd-trees{]}\end{tabular} &
  \greencheck &
  \crosscheck &
  FISTA &
  \crosscheck \\ \cline{2-12}

  %%%%
  %%% FACE 
\multicolumn{1}{|c|}{\cellcolor[HTML]{FFFC9E}} &
  \multicolumn{1}{l|}{FACE} &
  \multicolumn{1}{c|}{\citep{Poyiadzi20}} &
  \multicolumn{1}{c|}{\begin{tabular}[c]{@{}c@{}}C\\ {[}Tab / Img{]}\end{tabular}} &
  Yes \citep{FACE} &
  \crosscheck &
  \greencheck &
  \crosscheck &
  \greencheck &
  \greencheck &
  $\epsilon$ -graphs &
  \crosscheck \\ \cline{2-12} 
  
  %%%%
  %%% WEIGHTED COUNTERFACTUALS
\multicolumn{1}{|c|}{\cellcolor[HTML]{FFFC9E}} &
  \multicolumn{1}{l|}{Weighted Counterfactual} &
  \multicolumn{1}{c|}{\citep{Grath2018}} &
  \multicolumn{1}{c|}{\begin{tabular}[c]{@{}c@{}}C\\ {[}Tab{]}\end{tabular}} &
  No &
  \begin{tabular}[c]{@{}c@{}}\greencheck\\ {[}L$_1$-norm{]}\end{tabular} &
 \crosscheck &
  \greencheck &
  \crosscheck &
  \crosscheck &
  Gradient Descent &
  \crosscheck \\ \cline{2-12} 
  
  %%%%
  %%% TRUCE
\multicolumn{1}{|c|}{\cellcolor[HTML]{FFFC9E}} &
  \multicolumn{1}{l|}{TRUCE} &
  \multicolumn{1}{c|}{\citep{Laugel18,Laugel19,Laugel2019unjustified}} &
  \multicolumn{1}{c|}{\begin{tabular}[c]{@{}c@{}}C\\ {[}Tab / Txt / Img{]}\end{tabular}} &
  Yes \citep{TRUCE} &
  \begin{tabular}[c]{@{}c@{}}\greencheck\\ {[}L$_0$-norm{]}\end{tabular} &
  \crosscheck &
  \greencheck &
  \crosscheck &
  \crosscheck &
  Growing Spheres &
  \crosscheck \\ \cline{2-12} 
  
  %%%%
  %%% DICE 
\multicolumn{1}{|c|}{\multirow{-7}{*}{\cellcolor[HTML]{FFFC9E}\begin{tabular}[c]{@{}c@{}}Instance-Centric %
\end{tabular}}} &
  \multicolumn{1}{l|}{DICE} &
  \multicolumn{1}{c|}{\citep{Mothilal20}} &
  \multicolumn{1}{c|}{\begin{tabular}[c]{@{}c@{}}C\\ {[}Tab{]}\end{tabular}} &
  Yes \citep{DICE} &
  \begin{tabular}[c]{@{}c@{}}\greencheck\\ {[}L$_1$-norm{]}\end{tabular} &
  \crosscheck &
  \begin{tabular}[c]{@{}c@{}}\greencheck\\ {[}hinge loss{]}\end{tabular} &
  \greencheck &
  \greencheck &
  Gradient Descent &
  \crosscheck \\ 
  
  \hline
  \hline

    %%%%
  %%%% PROBABILITY-BASED
\multicolumn{1}{|c|}{\cellcolor[HTML]{FFCCC9}\begin{tabular}[c]{@{}c@{}}
\end{tabular}} &
  \multicolumn{1}{l|}{\begin{tabular}[c]{@{}l@{}}CRUDS\end{tabular}} &
  \multicolumn{1}{c|}{\citep{Downs2020crud}} &                                            % citation
  \multicolumn{1}{c|}{\begin{tabular}[c]{@{}c@{}}C\\ {[}Tab{]}\end{tabular}} &
  No &                                                                                    % source code
  \begin{tabular}[c]{@{}c@{}}\greencheck\\ {[}L$_2$-norm{]}\end{tabular} &                % proximity
  \greencheck &                                                                           % plausibility
  \begin{tabular}[c]{@{}c@{}}\crosscheck \\ {[}Variation Autoencoders{]} \end{tabular} &  % sparcity
  \greencheck &                                                                           % diversity
  \greencheck &                                                                           % feasibility
  \begin{tabular}[c]{@{}c@{}} - \end{tabular} &                  % optimisation
  \crosscheck \\ \cline{2-12}                                                                   % causal

 %%%
 %%% AReS
 \multicolumn{1}{|c|}{\cellcolor[HTML]{FFCCC9}\begin{tabular}[c]{@{}c@{}}
\end{tabular}} &
  \multicolumn{1}{l|}{\begin{tabular}[c]{@{}l@{}}AReS \\ {[}Global{]}\end{tabular}} &
  \multicolumn{1}{c|}{\citep{Rawal2020}} &                                                % citation
  \multicolumn{1}{c|}{\begin{tabular}[c]{@{}c@{}}C/R\\ {[}Tab/Txt{]}\end{tabular}} &
  No &                                                                                    % source code
  \begin{tabular}[c]{@{}c@{}}\greencheck\\ {[}{]}\end{tabular} &             % proximity
  \crosscheck &                                                                           % plausibility
  \begin{tabular}[c]{@{}c@{}}\crosscheck \\ {[} Probabilistic {[} \end{tabular} &  % sparcity
  \crosscheck &                                                                           % diversity
  \crosscheck &                                                                           % feasibility
  \begin{tabular}[c]{@{}c@{}} Maximum a Posterior \\ Estimate\end{tabular} &                      % optimisation
  \crosscheck \\ \cline{2-12}                                                                   % causal

 %%%
 %%% PRINCE
 \multicolumn{1}{|c|}{\cellcolor[HTML]{FFCCC9}\begin{tabular}[c]{@{}c@{}} 
\end{tabular}} &
  \multicolumn{1}{l|}{\begin{tabular}[c]{@{}l@{}}PRINCE\end{tabular}} &
  \multicolumn{1}{c|}{\citep{Ghazimatin20}} &                                                % citation
  \multicolumn{1}{c|}{\begin{tabular}[c]{@{}c@{}}C/R\\ {[}Tab/Txt{]}\end{tabular}} &
  Yes \citep{PRINCE} &                           % source code
  \begin{tabular}[c]{@{}c@{}}\greencheck\\ \end{tabular} &             % proximity
  \crosscheck &                                                                           % plausibility
  \begin{tabular}[c]{@{}c@{}}\crosscheck \\ {[} Random Walk {[} \end{tabular} &                   %  sparcity
  \crosscheck &                                                                           % diversity
  \crosscheck &                                                                           % feasibility
  \begin{tabular}[c]{@{}c@{}} PageRank \end{tabular} &                                    % optimisation
  \crosscheck \\ \cline{2-12}                                                                   % causal
 
  %%%
 %%% C-CHVAE
\multicolumn{1}{|c|}{\cellcolor[HTML]{FFCCC9}\begin{tabular}[c]{@{}c@{}} Probabilistic-Centric
\end{tabular}} &
  \multicolumn{1}{l|}{\begin{tabular}[c]{@{}l@{}}C-CHVAE\end{tabular}} &
  \multicolumn{1}{c|}{\citep{Pawelczyk2020CCHVAE,Pawelczyk2019}} &                        % citation
  \multicolumn{1}{c|}{\begin{tabular}[c]{@{}c@{}}C\\ {[}Tab{]}\end{tabular}} &
  Yes \citep{CCHVAE} &                           % source code
  \begin{tabular}[c]{@{}c@{}}\greencheck\\ \end{tabular} &             % proximity
  \crosscheck &                                                                           % plausibility
  \begin{tabular}[c]{@{}c@{}}\greencheck \\{[}Variation Autoencoders{]} \end{tabular} &   % sparcity
  \crosscheck &                                                                           % diversity
  \crosscheck &                                                                           % feasibility
  \begin{tabular}[c]{@{}c@{}} Integer Programming \\ Optimization \end{tabular} &                      % optimisation
  \crosscheck \\    \cline{2-12}                                             % causal

   %%%
 %%% ABELE
\multicolumn{1}{|c|}{\cellcolor[HTML]{FFCCC9}\begin{tabular}[c]{@{}c@{}}
\end{tabular}} &
  \multicolumn{1}{l|}{\begin{tabular}[c]{@{}l@{}}ABELE\end{tabular}} &
  \multicolumn{1}{c|}{\citep{Guidotti20}} &                        % citation
  \multicolumn{1}{c|}{\begin{tabular}[c]{@{}c@{}}C\\ {[}Img{]}\end{tabular}} &
  Yes \citep{ABELE} &                           % source code
  \begin{tabular}[c]{@{}c@{}}\greencheck\\ \end{tabular} &             % proximity
  \crosscheck &                                                                           % plausibility
  \begin{tabular}[c]{@{}c@{}}\greencheck \\{[}Variation Autoencoders{]} \end{tabular} &   % sparcity
  \crosscheck &                                                                           % diversity
  \crosscheck &                                                                           % feasibility
  \begin{tabular}[c]{@{}c@{}} - \end{tabular} &                      % optimisation
  \crosscheck \\    \cline{2-12}                                                  % causal
 
  %%% RECOURSE
\multicolumn{1}{|c|}{\cellcolor[HTML]{FFCCC9}\begin{tabular}[c]{@{}c@{}}
\end{tabular}} &
  \multicolumn{1}{l|}{\begin{tabular}[c]{@{}l@{}}RECOURSE\end{tabular}} &
  \multicolumn{1}{c|}{\citep{Karimi2020nips}} &                        % citation
  \multicolumn{1}{c|}{\begin{tabular}[c]{@{}c@{}}C\\ {[}Tab{]}\end{tabular}} &
  Yes \citep{RECOURSE} &                           % source code
  \begin{tabular}[c]{@{}c@{}}\greencheck\\ \end{tabular} &             % proximity
  \greencheck &                                                                           % plausibility
  \begin{tabular}[c]{@{}c@{}}\greencheck \\ {[}Variation Autoencoders{]} \end{tabular} &   % sparcity
  \greencheck &                                                                           % diversity
  \greencheck &                                                                           % feasibility
  \begin{tabular}[c]{@{}c@{}} Gradient Descent \end{tabular} &                      % optimisation
  \greencheck \\    \cline{2-12}  
  
  %%% MC-BRP
\multicolumn{1}{|c|}{\cellcolor[HTML]{FFCCC9}\begin{tabular}[c]{@{}c@{}}
\end{tabular}} &
  \multicolumn{1}{l|}{\begin{tabular}[c]{@{}l@{}}MC-BRP\end{tabular}} &
  \multicolumn{1}{c|}{\citep{Lucic2020}} &                        % citation
  \multicolumn{1}{c|}{\begin{tabular}[c]{@{}c@{}}R\\ {[}Tab{]}\end{tabular}} &
  Yes \citep{MCBRP} &                           % source code
  \begin{tabular}[c]{@{}c@{}}\greencheck\\ \end{tabular} &             % proximity
  \crosscheck &                                                                           % plausibility
  \begin{tabular}[c]{@{}c@{}}\greencheck \\ \end{tabular} &   % sparcity
  \greencheck &                                                                           % diversity
  \crosscheck &                                                                           % feasibility
  \begin{tabular}[c]{@{}c@{}} Monte Carlo \end{tabular} &                      % optimisation
  \crosscheck \\  
  
  \hline
  \hline
 
  %% CONSTRAINT
  %%%
  %%% GIC
  \multicolumn{1}{|c|}{\cellcolor[HTML]{FFCC67}} &
  \multicolumn{1}{l|}{GIC} &
  \multicolumn{1}{c|}{\citep{Lash2017}} &
  \multicolumn{1}{c|}{\begin{tabular}[c]{@{}c@{}}C\\ {[}Tab{]}\end{tabular}} &
  No &
  \greencheck &
  \crosscheck &
  \greencheck &
  \crosscheck &
  \crosscheck &
  \begin{tabular}[c]{@{}c@{}}Hill Climbing /\\ Genetic Algorithms\end{tabular} &
  \crosscheck \\ \cline{2-12} 
  
  %%%
  %%% MACE
\multicolumn{1}{|c|}{\cellcolor[HTML]{FFCC67}} &
  \multicolumn{1}{l|}{MACE} &
  \multicolumn{1}{c|}{\citep{Karimi2020mace}} &
  \multicolumn{1}{c|}{\begin{tabular}
  [c]{@{}c@{}}C\\ {[}Tab{]}\end{tabular}} &
  Yes \citep{MACE} &
  \begin{tabular}[c]{@{}c@{}}\greencheck\\ {[}L$_0$/L$_1$/L$_inf$-norm{]}\end{tabular} &
  \greencheck &
  \begin{tabular}[c]{@{}c@{}}\greencheck\\ {[}constraint satisfaction{]}\end{tabular} &
  \greencheck &
  \greencheck &
  SMT &
  \crosscheck \\ \cline{2-12}

  %%% 
  %%% COHERENT COUNTERFACTUALS
\multicolumn{1}{|c|}{\multirow{-3}{*}{\cellcolor[HTML]{FFCC67}\begin{tabular}[c]{@{}c@{}}Constraint-Centric
\end{tabular}}} &
  \multicolumn{1}{l|}{Coherent Counterfactuals} &
  \multicolumn{1}{c|}{\citep{Russell19}} &
  \multicolumn{1}{c|}{\begin{tabular}[c]{@{}c@{}}C /R\\ {[}Tab / Txt{]}\end{tabular}} &
  Yes \citep{COHERENT_COUNTERFACTUALS} &
  \begin{tabular}[c]{@{}c@{}}\greencheck\\ {[}L$_1$-norm{]}\end{tabular} &
  \greencheck &
  \begin{tabular}[c]{@{}c@{}}\greencheck\\ {[}mixed polytopes{]}\end{tabular} &
  \greencheck &
  \greencheck &
  \begin{tabular}[c]{@{}c@{}}Gurobi \\ Optimization\end{tabular} &
  \crosscheck \\ 
  
  \hline
  \hline
  
  %%%
  %%% MOCE
  \multicolumn{1}{|c|}{\cellcolor[HTML]{BCE8AD}} &
  \multicolumn{1}{l|}{MOCE} &
  \multicolumn{1}{c|}{\citep{Dandl2020}} &
  \multicolumn{1}{c|}{\begin{tabular}[c]{@{}c@{}}C\\ {[}Tab{]}\end{tabular}} &
  Yes \citep{MOCE} &
  \begin{tabular}[c]{@{}c@{}}\greencheck\\ {[}L$_0$-norm{]}\end{tabular} &
  \crosscheck &
  \begin{tabular}[c]{@{}c@{}}\greencheck\\ {[} min feature changes{]}\end{tabular} &
  \crosscheck &
  \crosscheck &
  NSGA-II &
  \crosscheck \\ \cline{2-12} 
  
%%%
%%% CERTIFAI
\multicolumn{1}{|c|}{\cellcolor[HTML]{BCE8AD}} &
  \multicolumn{1}{l|}{CERTIFAI} &
  \multicolumn{1}{c|}{\citep{Sharma19}} &
  \multicolumn{1}{c|}{\begin{tabular}[c]{@{}c@{}}C\\ {[}Tab / Img {]}\end{tabular}} &
  Yes \citep{CERTIFAI} &
  \begin{tabular}[c]{@{}c@{}}\greencheck\\ {[}L$_1$-norm / SSIM{]}\end{tabular} &
  \crosscheck &
  \crosscheck &
  \begin{tabular}[c]{@{}c@{}}\greencheck\\ {[}mutations{]}\end{tabular} &
  \crosscheck &
  Fitness &
  \crosscheck \\ \cline{2-12}

 %%%
 %%% LORE 
\multicolumn{1}{|c|}{\multirow{-3}{*}{\cellcolor[HTML]{BCE8AD}\begin{tabular}[c]{@{}c@{}}Genetic-Centric 
\end{tabular}}} &
  \multicolumn{1}{l|}{LORE} &
  \multicolumn{1}{c|}{\citep{Guidotti18}} &
  \multicolumn{1}{c|}{\begin{tabular}[c]{@{}c@{}}C\\ {[}Tab{]}\end{tabular}} &
  Yes \citep{LORE} &
  \begin{tabular}[c]{@{}c@{}}\greencheck\\ {[}L$_2$-norm / Match{]}\end{tabular} &
  \crosscheck &
  \crosscheck &
  \begin{tabular}[c]{@{}c@{}}\greencheck\\ {[}mutations{]}\end{tabular} &
  \crosscheck &
  \begin{tabular}[c]{@{}c@{}}DecisionTree \\ Model\end{tabular} &
  \crosscheck \\ 
  
  \hline
  \hline
  
  %%%%
  %%%% LIME-C
\multicolumn{1}{|c|}{\cellcolor[HTML]{DAE8FC}} &
  \multicolumn{1}{l|}{LIME-C} &
  \multicolumn{1}{c|}{\citep{Ramon19,Ramon20}} &
  \multicolumn{1}{c|}{\begin{tabular}[c]{@{}c@{}}C / R\\ {[}Tab / Txt / Img{]}\end{tabular}} &
  Yes \citep{LIMEC} &
  \crosscheck &
  \crosscheck &
  \crosscheck &
  \crosscheck &
  \crosscheck &
  \begin{tabular}[c]{@{}c@{}}Additive Feature\\ Attribution\end{tabular} &
  \crosscheck \\ \cline{2-12}

  %%%%
  %%%% SED-C
\multicolumn{1}{|c|}{\cellcolor[HTML]{DAE8FC}} &
  \multicolumn{1}{l|}{SED-C} &
  \multicolumn{1}{c|}{\citep{Martens14}} &
  \multicolumn{1}{c|}{\begin{tabular}[c]{@{}c@{}}C\\ {[}Txt{]}\end{tabular}} &
  Yes \citep{SEDC} &
  \begin{tabular}[c]{@{}c@{}}\crosscheck\\ {[}cosine similarity{]}\end{tabular} &
  \crosscheck &
  \crosscheck &
  \crosscheck &
  \crosscheck &
  - &
  \crosscheck \\ \cline{2-12} 
  
  %%%%
  %%%% CLEAR
\multicolumn{1}{|c|}{\multirow{-3}{*}{\cellcolor[HTML]{DAE8FC}\begin{tabular}[c]{@{}c@{}}Regression-Centric
\end{tabular}}} &
  \multicolumn{1}{l|}{CLEAR} &
  \multicolumn{1}{c|}{\citep{White19}} &
  \multicolumn{1}{c|}{\begin{tabular}[c]{@{}c@{}}C\\ {[}Tab{]}\end{tabular}} &
  Yes\citep{CLEAR} &
  \begin{tabular}[c]{@{}c@{}}\greencheck\\ {[}L$_2$-norm{]}\end{tabular} &
  \crosscheck &
  \begin{tabular}[c]{@{}c@{}}\greencheck\\ {[} min feature changes{]}\end{tabular} &
  \crosscheck &
  \crosscheck &
  Regression &
  \crosscheck \\ 
  
  \hline
  \hline

  %%%%
  %%%% SHAP-C
\multicolumn{1}{|c|}{\cellcolor[HTML]{CBCEFB}} &
  \multicolumn{1}{l|}{SHAP-C} &
  \multicolumn{1}{c|}{\citep{Ramon19,Ramon20}} &
  \multicolumn{1}{c|}{\begin{tabular}[c]{@{}c@{}}C / R\\ {[}Tab / Txt / Img{]}\end{tabular}} &
  Yes \citep{SHAPC} &
  \crosscheck &
  \crosscheck &
  \crosscheck &
  \crosscheck &
  \crosscheck &
  Shapley Values &
  \crosscheck \\ \cline{2-12}

  %%%%
  %%%% SHAP-CC
\multicolumn{1}{|c|}{\multirow{-2}{*}{\cellcolor[HTML]{CBCEFB}\begin{tabular}[c]{@{}c@{}}Game Theory\\ Centric %Approaches JAJ
\end{tabular}}} &
  \multicolumn{1}{l|}{SHAP-CC} &
  \multicolumn{1}{c|}{\citep{Rathi2019}} &
  \multicolumn{1}{c|}{\begin{tabular}[c]{@{}c@{}}C / R\\ {[}Tab{]}\end{tabular}} &
  No &
  \crosscheck &
  \crosscheck &
  \crosscheck &
  \crosscheck &
  \crosscheck &
  Shapley Values &
  \crosscheck \\ 
  
  \hline
  \hline
  
  %%%%
  %%%% CBR
\multicolumn{1}{|c|}{\cellcolor[HTML]{9698ED}\begin{tabular}[c]{@{}c@{}}Case Based Reasoning 
\end{tabular}} &
  \multicolumn{1}{l|}{\begin{tabular}[c]{@{}l@{}}CBR for Good\\ Counterfactuals\end{tabular}} &
  \multicolumn{1}{c|}{\citep{Keane2020}} &
  \multicolumn{1}{c|}{\begin{tabular}[c]{@{}c@{}}C\\ {[}Tab / Txt{]}\end{tabular}} &
  No &
  \begin{tabular}[c]{@{}c@{}}\greencheck\\ {[}L$_1$-norm{]}\end{tabular} &
  \greencheck &
  \begin{tabular}[c]{@{}c@{}}\greencheck\\ {[}counterfactual potential{]}\end{tabular} &
  \greencheck &
  \greencheck &
  \begin{tabular}[c]{@{}c@{}}Nearest Unlikely\\ Neighbour\end{tabular} &
  \crosscheck \\ 

  \hline
\end{tabular}
}
\caption{Classification of collected model-agnostic counterfactual algorithms for XAI based on different properties, theoretical backgrounds and applications.}
\label{tab:algo_summary}
\end{table}

\section{Counterfactual Approaches to Explainable AI: Applications}\label{sec:counterfactuals_applications}

This work is motivated by \citet{Holzinger19} hypothesis, which states that for a system to provide understandable human explanations, the user needs to achieve a specified level of causal understanding with effectiveness, efficiency, and satisfaction in a specified context of use~\citep{Holzinger21,Holzinger20icom,Holzinger17medicaldomain,Holzinger18,Holzinger20, Xu2020Causality}. One way to achieve this causal understanding is through counterfactuals.

One of the main areas that showed a need for counterfactual explanations is in medical decision support systems. As pointed by \citet{Holzinger19interactive}, medical decision-making faces several challenges ranging from small labelled datasets to the integration, fusion, and mapping of heterogeneous data to appropriate visualisations \citep{Holzinger14}. Structured causal models that provide explanatory factors of the data could be used to support medical experts. However, learning causal relationships from observational data is a very difficult problem~\citep{Zhao19,Peters19}.

XAI is very relevant for industries \citep{Rehse2019}. Another area of application of counterfactuals in XAI that is highly mentioned in the literature is loan credit evaluation. 
\citet{Grath2018} developed a model-agnostic counterfactual explainer with an interface that uses weights generated from feature importance to generate more compact and intelligible counterfactual explanations for end users. \citet{Lucic2020} also developed a model-agnostic counterfactual explanation in the context of a challenge from the finance industry’s interest in exploring algorithmic explanations~\citep{FICO2017}.

Recently, interfaces that generate counterfactuals as explanations have been proposed in the literature. The "What-IF tool" \citep{Wexler2019} is an open-source application that allows practitioners to probe, visualize, and analyze machine learning systems with minimal coding. It also enables the user to investigate decision boundaries and explore how general changes to data points affect the prediction. 

\textcolor{blue}{ViCE (Visual Counterfactual Explanations for Machine Learning Models)~\citep{Gomez2020} is an interactive visual analytics tool that generates instance-centric counterfactual explanations to contextualize and evaluate model decisions in a home equity line of credit scenario. ViCE highlights counterfactual explanations and equips users with interactive methods to explore both the data and the model.}

\textcolor{blue}{DECE (Decision Explorer with Counterfactual Explanations for Machine Learning Models)~\citep{Cheng20dece} is another example of an interactive visual analytics tool that generates counterfactuals at an instance and subgroup levels. The main difference to ViCE is that DECE allows users to interact with the counterfactuals in order to find more actionable counterfactuals that suit their needs. DECE showed effectiveness in supporting decision exploration tasks and instance explanations.}

%\citet{Nori2019interpret}
%As Zhang et al. (2020) mention, more work is needed on whether causal information leads to better decisions. Many unstudied factors may contribute to this diverse literature on the topic ranging from human cognitive bias to the way the explanations are presented in the interface (supporting interactivity or not).   

\section{Towards Causability: Opportunities for Research}\label{sec:causability}

\textcolor{blue}{Although the demand for providing XAI systems that promote the causability, the literature is very scarce in this aspect. We only found one recent article that proposed an explanation framework (FATE) based on causability \cite{Shin2021}. This framework focuses on human interaction, and the authors used the system causability scale proposed by \cite{Holzinger19} to validate the effectiveness of their system's explanations.}

\citet{Shin2021} highlight that causability represents the quality of explanations and emphasize that it is an antecedent role to explainability. Furthermore, they found that properties such as transparency, fairness, and accuracy, play a critical role in improving user trust in the explanations. In general, this framework is a guideline for developing user-centred interface design from the perspective of user interaction and examines the effects of explainability in terms of user trust. This framework does not refer to XAI algorithms underpinned by a theory of causality, neither on how to achieve causability from such mathematical constructs. In the next section, we provide a set of conditions that we find are crucial elements for an XAI system to promote causability. However, FATE causability system is not underpinned by any formal theory of causality, and the causability metrics applied in this work focused on the interaction of the human with the system.

\citet{Holzinger19} proposed a theoretical framework with a set of guidelines to promote causability in XAI systems in the medical domain. One of the policies put forward is in creating new visualization techniques that can be trainable by medical experts, as the specialists can survey the underlying explanatory factors of the data. Another point is to formalize a structural causal model of human decision-making and delineating features in the model. \citet{Holzinger20icom} argue that a human-AI interface with counterfactual explanations will help achieve causability. An open research opportunity is to extend human-AI explainable interfaces with causability by allowing a domain expert to interact and ask “what-if” questions (counterfactuals). This will enable the user to gain insights into the underlying explanatory factors of the predictions. \citet{Holzinger20} propose a system causability scale framework as an evaluation tool for causability in XAI systems.

We conclude our systematic literature review by highlighting what properties should an XAI system have to promote causability. We find that the process of generating explanations that are human-understandable needs to go beyond the minimisation of some loss function as proposed by the majority of the algorithms in the literature. Explainability is a property that implies the generation of human mental representations that can provide some degree of human-understandability of the system and, consequently, allow users to trust it. As \citet{Guidotti18} stated, explainability is the ability to present interpretations in a meaningful and effective way to a human user. 
We argue that for a system do be both explainable and promote causability, then it cannot be resumed to a minimisation optimisation problem. Doing so would imply a simplistic and objective explanation process that needs to be necessarily human-centric to achieve human understandability~\citep{Confalonieri21}. We argue that, for a system to promote causability, the following properties should be satisfied:

\subsection{The Main Characteristics of a Causability System}

\begin{itemize}

\item \textbf{Causality.} The analysis we conducted revealed that current model-agnostic explainable AI algorithms lack a foundation on a formal theory of causality. Causal explanations are a crucial missing ingredient for opening the black box to render it understandable to
human decision-makers since knowing about the cause/effect relationships of variables can promote human understandability. We argue that causal approaches should be emphasised in XAI to promote a higher degree of interpretability to its users and causability, \textcolor{blue}{although some authors challenge this notion. \citet{Zheng2020} found that providing causal information to human users in some tasks, resulted in poor decision-making. \citet{Zheng2020} conducted studies to investigate whether presenting causal explanations to a user would lead to better decision-making. Their results were mixed. They found that if the user has prior and existing domain knowledge, then presenting causal information did not improve the decision-making quality of the user. On the other hand, if the user did not have any prior knowledge or beliefs about a specific task, then causal information enabled the user to make better decisions. More work is needed on whether causal information leads to better decisions. Many unstudied factors may contribute to this diverse literature on the topic ranging from human cognitive bias to the way the explanations are presented in the interface (supporting interactivity or not).}  %% explain here nad 
%\citep{Chen2020CausalML}
%\citep{Mothilal20b}
\item \textbf{Counterfactual.} Explanations generated by a causability system needs to be counterfactual. Cognitive scientists agree that counterfactual reasoning is a crucial ingredient in learning and a key for explaining adaptive behaviour in a changing environment~\citep{Paik2014}. Counterfactual reasoning induces mental representations of an event that happened and representations of some other event alternative to it~\citep{Stepin2021}. \textcolor{blue}{ It has been recognised in the literature that counterfactuals tend to help humans make causal judgments \citep{Gerstenberg17}. Additionally, humans tend to think in a cause/effect way, but not in a strict probabilistic sense \citep{Goldvarg01}}. It follows that for a machine to achieve a certain degree of human intelligence, explainability systems need to provide counterfactual explanations. Additionally, for a system to achieve causability, the counterfactual explanations need to be underpinned by a formal theory of causality~\citep{Holzinger19,Holzinger20}. Properties to generate good counterfactuals, such as diversity, feasibility, and plausibility, should also be considered to increase the level of human understanding. 

\item \textbf{Human-Centric.} Explanations need to be adapted to the information needs of different users. For instance, in medical decision-making, a doctor is interested in certain aspects of an explanation, while a general user is interested in other types of information. Adapting the information for the type of user is a crucial and challenging point currently missing in XAI literature. There is the need to bring the human user back to the optimisation process with human-in-the-loop strategies \citep{Holzinger16,Holzinger19interactive} containing contextual knowledge and domain-specific information. This interactive process can promote causability since it will allow the user to create mental representations of the counterfactual explanations in a symbiotic process between the human and the counterfactual generation process. 

\item \textbf{Inference.} To promote the system's user understandability, we argue that a causability framework should be equipped with causal inference mechanisms to interact with the system and ask queries to the generated explanations. Queries such as "given that I know my patient has a fever, what changes this information induces in the explanation?".  This type of interaction can be highly engaging for the user and promote more transparency in the system. It can enable more human-centric understandability of the system since the user asks questions (performs inference) over variables of interest.

\item \textbf{Semantic annotations.} One of the major challenges in XAI and a current open research problem is to convert the sub-symbolic information extracted from the black-box into human-understandable explanations. Incorporating semantic contextual knowledge and domain-specific information are crucial ingredients that are currently missing in XAI. We argue that story models and narratives are two important properties that need to be considered to generate human-understandable and human-centric explanations. Story models and narratives can promote higher degrees of believability in the system \citep{Yale13} and consequently achieve causability. 

\end{itemize}

\section{Answers to Research Questions}\label{sec:rq}

This section summarises the key points presented throughout this work by providing answers to the research questions that guided our research.

\subsection{RQ1 \& RQ2: What are the main theoretical approaches and algorithms for counterfactuals in XAI?}

Our systematic literature review revealed many different counterfactual algorithms proposed in the literature. We were able to identify key elements shared by these algorithms based on how the optimisation problem was framed and by considering the counterfactual generation process. We classified the existing model-agnostic-XAI by their \textit{"master theoretical algorithm"} from which each algorithm derived:

\begin{itemize}
	\item  \textbf{Instance-Centric.} These approaches are based on random feature permutations and on finding counterfactuals closed to the original instance by some distance function. These approaches are relatively straightforward to implement. However, the generated counterfactuals are susceptible to fail the plausibility and the diversity properties, although some algorithms incorporate mechanisms to overcome this issue. Examples of algorithms that fall in this category are WatcherCF~\citep{Wachter18}, prototype counterfactuals \citep{Looveren2019}, weighted counterfactuals~\citep{Grath2018},FACE~\citep{Poyiadzi20}, DiCE~\citep{Mothilal20}, and TRUCE~\citep{Laugel19}.

\item \textbf{Constraint-Centric.} 
These approaches are modelled as a constraint satisfaction problem. Counterfactuals based on these approaches can easily satisfy different properties such as diversity and plausibility. Examples of algorithms that fall in this category are MACE~\citep{Karimi2020mace}, GIC~\citep{Lash2017}, and Coherent Counterfactuals~\citep{Russell19}.

\item \textbf{Genetic-Centric.}
These approaches generate counterfactuals using the principles of genetic algorithms. Due to genetic principles such as mutation or crossover, these approaches can satisfy counterfactual properties such as diversity and plausibility. Examples of algorithms that fall in this category are CERTIFAI~\citep{Sharma19}, MOCE~\citep{Dandl2020}, and LORE~\citep{Guidotti18}.

\item \textbf{Regression-Centric.} 
These approaches have LIME as their underlying framework, and they use linear regression to fit a set of permutated features. Counterfactuals based on these approaches have difficulties satisfying several properties such as plausibility and diversity. Examples of algorithms that fall in this category are LIME-C~\citep{Ramon20}, SED-C~\citep{Martens14}, and CLEAR~\citep{White19}.

\item \textbf{Game Theory Centric.} 
These approaches have SHAP as their underlying framework, and they use Shapley values to determine the local feature relevance. Counterfactuals based on these approaches also have difficulties satisfying several properties such as plausibility and diversity.  Examples of algorithms that fall in this category are SHAP-C~\citep{Ramon20}, and SHAP-CC~\citep{Rathi2019}.

\item \textbf{Case-Based Reasoning.}
These approaches are inspired in the case-based reasoning paradigm of artificial intelligence and cognitive science. Since they store in-memory examples of good counterfactuals, these approaches tend to satisfy  different counterfactual properties, such as plausibility and diversity. An examples of an algorithms that fall in this category is CBR Explanations by~\citet{Keane2020}.

\textcolor{blue}{\item \textbf{Probabilistic-Centric.}}
 \textcolor{blue}{These approaches mainly use probabilistic models to find the nearest counterfactuals. Approaches such as \textit{recourse}~\citep{Karimi2020nips} have the potential to generate causal counterfactuals based on the causality framework proposed by \citet{Pearl09}. However, as the authors acknowledge, it is challenging to learn causal relationships from observational data without introducing assumptions in the causal model.}

\end{itemize}

This research suggests that current model-agnostic counterfactual algorithms for explainable AI are not grounded on a causal theoretical formalism and, consequently, might not promote causability to a human decision-maker. Our findings show that the explanations derived from most of the model-agnostic algorithms in the literature provide spurious correlations rather than cause/effects relationships, leading to sub-optimal, erroneous, or even biased explanations. This opens the door to new research directions on incorporating formal causal theories of causation in XAI. \textcolor{blue}{The closest work that we found that meets this goal is the \textit{Recouse} algorithm~\citep{Karimi2020nips}, however, research is still needed to investigate the extraction of structured causal models from observational data.} 

\textcolor{blue}{There are also novel model-agnostic approaches proposed in the literature of XAI based on probabilistic graphical models. For instance, \citet{Moreira21dss} proposed to learn a local Bayesian network that enables the user to see which features are correlated (or conditional independent from) the class variable. They found four different rules that can measure the degree of confidence of the interpretable model over the explanations and provide specific recommendations for the user. However, this model is not causal, and further research is needed to understand if such structures can be mapped into structured causal models.} 

\subsection{RQ3: What are the sufficient and necessary conditions for a system to promote causability (Applications)?}

The main purpose of this research work is to highlight some properties that find relevant and necessary for causability systems. We proposed the following properties. 

\begin{itemize}
    \item Explanations need to be grounded on a structured and formal theory of Causality. This will enable the usage of a full framework of algorithms of causal discovery that have been proposed throughout the years~\citep{Peters17}.
    
    \item Explanation algorithms need to be computed in the form of Counterfactuals. Due to the evidence from cognitive science and social sciences, counterfactuals are among the best approaches to promote human understandability and interpretability~\citep{Miller2019}, \textcolor{blue}{although some authors challenge this~\citep{Zheng2020}.}
    
    \item Explanations need to be Human-Centric. Explanations need to be specific to the user's needs: a medical doctor will be interested in different explanations from a standard user.
    
    \item The user should be able to interact with the generated explanations. The interaction with explanations can help the user increase the levels of understandability and interpretability of the internal workings of the XAI algorithm. Probabilistic inference is a promising tool to provide answers to users' questions regarding explanations.
    
    \item Explainable AI systems need to be complemented with semantic annotations of features and domain knowledge. To achieve explainability, contextual knowledge and domain-specific information need to be included in the system.
\end{itemize}

\subsection{ RQ4: What are the pressing challenges and research opportunities in XAI systems that promote Causability?}

This literature review enabled us to understand the current pressing challenges and opportunities involved in creating XAI models that promote causability. We identified the following research opportunities that can be used for future research in the area.

\begin{itemize}
    \item \textbf{Causal Theories for XAI.} 
    \citet{Pear19} argues that causal reasoning is indispensable for machine learning to reach the human-level artificial intelligence since it is the primary mechanism of humans to be aware of the world. As a result, the causal methodology gradually becomes a vitally important component in explainable and interpretable machine learning. However, most current interpretability techniques pay attention to solving the correlation statistic rather than the causation. Therefore, causal approaches should be emphasized to achieve a higher degree of interpretability. The reason why causal approaches for XAI are scarce is that finding causal relationships from observational data is very hard and still an open research question \citep{Zhao19}.
    
    \item \textbf{Standardized Evaluation Metrics for XAI.} 
    The field of metrics for XAI is also a topic that needs development. Measures such as stability or fidelity \citep{Velmurugan2020} are not very clear for counterfactuals \citep{Carvalho19}. Ultimately, XAI metrics should be able to answer the following question: \textit{how does one know whether the explanation works and the user has achieved a pragmatic understanding of the AI?}~\citep{hoffman2019metrics}  We highlight that one research concern in XAI should be to develop generalised and standardised evaluation protocols for XAI in different levels: Objective Level (user-free), Functional Level (functionality-oriented), and User Level (human-centric). The main challenges consist in deriving standardised protocols that could fit so many algorithms underpinned by different \textit{master theoretical approaches} and at so many different levels, although some interesting works have already been proposed in terms of causability~\citep{Holzinger20icom,Holzinger20,Holzinger21}.

    \item \textbf{Intelligent Interfaces for Causability in XAI.} 
    XAI's basilar applications lie at the core of Intelligent User Interfaces (IUIs). Rather than generating explanations as linear symbolic sequences, graphical interfaces enable people to visually explore ML systems to understand how they perform over different stimuli. The What-If tool~\citep{Wexler2019} provides an excellent example, enabling people to visualize model behavior across multiple models and subsets of input data and for different ML fairness metrics. Such visual techniques leverage the human visual channel's high bandwidth to explore probabilistic inference, allowing humans to interact with explanations while recommending different descriptions. Taking advantage of the innate human ability to spot patterns, these methods can provide better quality answers than purely automatic approaches. 
    
    In a related direction~\citet{Waa2021} propose a framework that considers users' experience and reactions to explanations and evaluates these effects in terms of understanding, persuasive power, and task performance. This user-centric approach is crucial to assess assumptions and intuitions to yield more effective explanations effectively.

    Recent Intelligent Exploration Interfaces focus on making explanations accessible to non-expert users to interpret the underlying models better. 
    \citet{hoque2021} argue that predictive and interactive models based on causality are inherently interpretable and self-contained. They developed Outcome Explorer, a causality-guided interactive interface that allows experts and non-experts alike to acquire a comprehensive understanding of the models.

\end{itemize}

\section{Conclusion}\label{sec:conclusion}

We conducted a systematic literature review to determine the modern theories underpinning model-agnostic counterfactual algorithms for XAI and analyse if any existing algorithms can promote causability. We extended the current literature by proposing a new taxonomy for model-agnostic counterfactuals based on six approaches: instance-centric, constraint-centric, genetic- centric, regression-centric, game theory centric, case-based reasoning centric, and probabilistic-centric. Our research also showed that model-agnostic counterfactuals are not based on a formal and structured theory of causality as proposed by~\citep{Pearl09}. For that reason, we argue that these systems cannot promote a causal understanding to the user without the risk of the explanations being biased, sub-optimal, or even erroneous. Current systems determine relationships between features through correlation rather than causation. 

We conclude this survey by highlighting new key points to promote causability in XAI systems, which derive from formal theories of causality such as inference, counterfactuals, and probabilistic graphical models. 
Causal models are a new research area, bursting with exciting new research challenges and opportunities for XAI approaches grounded on probabilistic theories of causality and graphical models. Indeed this field is highly relevant to Intelligent User Interfaces (IUIs)~\citep{IUI2020} by its very nature, both in terms of content generation engines and user interface architecture. Therefore, more than a contraption powered by robust and effective causal models, XAI can be seen as a cornerstone for next-generation IUIs. This can only be achieved by marrying sound explanations delivered by fluid storytelling to persuasive and articulate argumentation and a harmonious combination of different interaction modalities. These will usher in powerful engines of persuasion, ultimately leading to the rhetoric of causability. 

\section{Acknowledgement}

This work was partially supported by Portuguese government national
funds through FCT, Funda\c{c}\~{a}o para a Ci\^{e}ncia e a Tecnologia, under project UIDB/50021/2020.

\textcolor{blue}{This work was also partially supported by Queensland University of Technology (QUT) Centre for Data Science First Byte Funding Program and by QUT's Women in Research Grant Scheme.}

%\bibliography{my_works,Literature}

\end{document}